\documentclass{article} 
\usepackage{iclr2027_conference,times}
\iclrfinalcopy

\usepackage{amsmath,amsfonts,bm}

\def\eqref#1{equation~\ref{#1}}

\def\1{\bm{1}}

\DeclareMathAlphabet{\mathsfit}{\encodingdefault}{\sfdefault}{m}{sl}
\SetMathAlphabet{\mathsfit}{bold}{\encodingdefault}{\sfdefault}{bx}{n}

\usepackage[hidelinks]{hyperref}
\usepackage{url}
\usepackage{amsthm}
\usepackage{graphicx} 
\usepackage{booktabs}
\usepackage{subcaption}

\newtheorem{theorem}{Theorem}[section]
\newtheorem{lemma}[theorem]{Lemma}

\newtheorem{corollary}[theorem]{Corollary}

\theoremstyle{definition}
\newtheorem{definition}[theorem]{Definition}

\theoremstyle{remark}

\title{Where Latents Belong and How They Flow: Joint Identification of Structural Roles
\par\vspace{0.2in}}

\author{
  \shortstack[c]{
    \textbf{Xintong Li} \\
    {\normalfont CMU} \\
    {\normalfont\texttt{lixintong@cmu.edu}}
  }
  \And
  \shortstack[c]{
    \textbf{Kun Zhang} \\
    {\normalfont CMU, MBZUAI} \\
    {\normalfont\texttt{kunz1@cmu.edu}}
  }
  \And
  \shortstack[c]{
    \textbf{Yujia Zheng} \\
    {\normalfont UIUC} \\
    {\normalfont\texttt{yujiaz@illinois.edu}}
  }
}

\begin{document}

\maketitle
\fancyhead{}
\renewcommand{\headrulewidth}{0.4pt}

\begin{abstract}
Latent factors in dynamical systems can play different roles across observations and time. For example, an idea in a group meeting may be expressed by only some participants, inherit information from earlier discussions, and shape what is said next. 
We formalize this notion as the \emph{structural role} of a latent factor, describing where it is expressed, whether it receives influence from the past, and whether it transmits influence to the future. 
The main challenge is correspondence, since the three structures may be identified under different permutations, while a structural role is defined only when they are aligned to the same latent coordinate.
We show that, under suitable structural and variability conditions, all three roles are jointly identifiable under a common permutation at each time point. Guided by this theory, we develop \textsc{RoleTrace}, a framework that traces structural roles in a shared latent coordinate system. Synthetic experiments validate structural recovery, while experiments on multiparty meetings examine the identified roles in real sequential interactions.
\end{abstract}


\section{Introduction}

\begin{figure}[t]
 \centering
 \includegraphics[width=\textwidth]{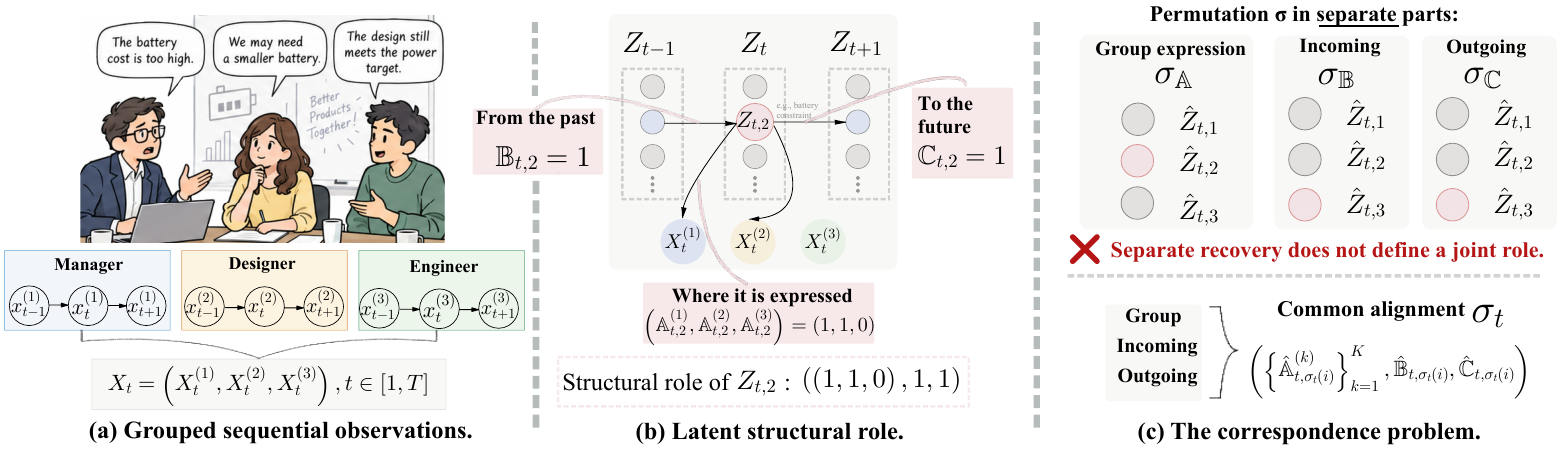}
 \caption{Example of latent structural roles and their alignment. For observation $X_t$ and its latent $Z_{t,2}$: $\mathbb A_{t,2}^{(1)}=1$ means that $Z_{t,2}$ is expressed in group~1, $\mathbb B_{t,2}=1$ indicates incoming influence, and $\mathbb C_{t,2}=1$ indicates outgoing influence. Joint identification requires a common permutation $\sigma_t$.}
 \label{fig:intro_overview}
 \vspace{-3pt}
\end{figure}

Understanding how hidden factors are organized and evolve is fundamental to explaining complex systems from their observations~\citep{pmlr-v97-locatello19a,scholkopf2021toward}. In many dynamic systems, these hidden factors can be expressed in different parts of the system and can continue to influence what happens next.
Consider a group meeting in Figure~\ref{fig:intro_overview} (a), where participants discuss different aspects of a project. We observe their remarks over time, while the latent factors shaping the discussion remain hidden. An underlying standpoint may be expressed in some participants' remarks, inherit influence from earlier discussion, and shape later decisions. Understanding such a factor therefore involves where it is expressed, what influence it receives from the past, and whether its influence continues into the future.
The same pattern appears beyond meetings, including multimodal streams~\citep{DBLP:journals/corr/abs-1806-06176,10.1145/3552487.3556435}, distributed sensing~\citep{4434303,7997940}, biological systems~\citep{velten2022identifying,3692070.3693197,pmlr-v267-liu25bh}, and other settings~\citep{wu2023iclr-slotformer,li2026spatio} where an evolving hidden process is observed through several meaningful parts of a system.
This raises a general question: 
\begin{center}
    \emph{Can we identify how hidden factors participate in an evolving system?}
\end{center}

These systems contain two complementary forms of structure: observations are structured across groups, while latent states are dependent across time. 
The group structure characterizes where each latent factor is expressed. At each time point, the groups may correspond to participants in a meeting, modalities in a multimodal system or regions in a spatial process. For example, in Figure~\ref{fig:intro_overview} (b), the latent $Z_{t,2}$ is expressed in the observations of the Manager $X_t^{(1)}$ and Designer $X_t^{(2)}$, but not the Engineer $X_t^{(3)}$, giving $\left(\mathbb A_{t,2}^{(1)},\mathbb A_{t,2}^{(2)},\mathbb A_{t,2}^{(3)}\right)=(1,1,0)$.
The temporal structure characterizes its incoming and outgoing influence. $Z_{t,2}$ receives influence from the previous state and affects the next, corresponding to $\mathbb B_{t,2}=1$ and $\mathbb C_{t,2}=1$.
Together, these attributes give $((1,1,0),1,1)$, which we call the \emph{structural role} of a latent variable. 

This formulation extends the usual target of latent variable identification. 
Independent component analysis (ICA) studies the recovery of latent sources from observed mixtures~\citep{COMON1994287,d243f4c2f46949b29cf034348d84c11f}.
Nonlinear ICA extends this setting, with identifiability established under additional structures such as auxiliary variables~\citep{pmlr-v89-hyvarinen19a,pmlr-v108-khemakhem20a}, restricted function classes~\citep{BucBesSch22, lachapelle2023additive}, or structural sparsity~\citep{3600270.3601464,3666122.3666707}. More recent work~\citep{zheng2025thought} further identifies the incidence structure between latent factors and observation groups.
For sequential data, temporal causal representation learning exploits temporal dependence to identify latent processes and their causal relations under assumptions on temporal variation~\citep{yao2022learning, 3600270.3602191, 3666122.3666477}, historical context~\citep{3692070.3692351}, or sparse latent dynamics~\citep{li2025identification}. Specifically,~\citep{li2025identification2} further allows the active latent support to vary over time, but the identified temporal support corresponds to our incoming role $\mathbb B$, with observation and outgoing activity coupled into the support definition.
Existing work therefore identifies latent components, observation-side structure, or temporal causal structure, while the complete role of a latent remains unresolved. 

The main challenge is establishing correspondence among these structural views. Latent coordinates have no intrinsic labels, so each structure is generally identified only up to permutation. As illustrated in Figure~\ref{fig:intro_overview}(c), separately recovering group expression, incoming influence, and outgoing influence under $\sigma_A$, $\sigma_B$, and $\sigma_C$ may assign the same true latent to different learned coordinates. 
Joint identification therefore requires a common permutation $\sigma_t$ that aligns all three structural views simultaneously:
$\left(\{\widehat{\mathbb A}_{t,\sigma_t(i)}^{(k)}\}_{k=1}^K, \widehat{\mathbb B}_{t,\sigma_t(i)}, \widehat{\mathbb C}_{t,\sigma_t(i)} \right) = \left( \{\mathbb A_{t,i}^{(k)}\}_{k=1}^K, \mathbb B_{t,i}, \mathbb C_{t,i} \right).$
Under this common alignment, the recovered attributes define the structural role of the same latent coordinate.

Our theory provides guarantees for recovering the complete structural role of each latent coordinate from grouped sequential observations. 
Variation in the grouped observation map reveals which latent directions are expressed in each group, while the row and column supports of the transition cross Hessian reveal which coordinates receive and transmit delayed influence. Since all three views constrain the same latent reparameterization at time $t$, their correspondence is recovered through a common permutation $\sigma_t$, yielding joint identification of $\mathbb A$, $\mathbb B$, and $\mathbb C$.
Guided by the theory, we develop \textsc{RoleTrace}, which learns a latent representation and its transition density in the same coordinate system. The decoder Jacobian estimates group expression, while derivatives of the transition density characterize temporal roles.
Our contributions are threefold:
\textbf{Conceptually,} we formulate structural role identification for grouped sequential observations, characterizing each latent coordinate by its group and temporal roles. \textbf{Theoretically,} we establish identifiability of all three structural roles under a common permutation, which enables recovery of their correspondence at each latent coordinate. \textbf{Practically,} we develop \textsc{RoleTrace}, a theory guided estimator of structural roles in a shared latent coordinate system, and evaluate it on both synthetic and real-world data.




\section{Preliminaries}
\label{sec:preliminaries}

In this section, we formulate the temporal data-generating process (DGP), introduce the structural support objects used in our analysis, and define the latent roles that will be identified in Section~\ref{sec:theory}. 

\textbf{Data-generating process.} For a positive integer $n$, let $[n]:=\{1,\ldots,n\}$. For time horizon $T$ and latent dimension $N$, we consider a latent sequence $Z_t=(Z_{t,1},\ldots,Z_{t,N})\in\mathcal Z_t\subseteq\mathbb R^N$, $t\in[T]$. We observe $X_{1:T}:=(X_1,\ldots,X_T)$, where $X_t=(X_t^{(1)},\ldots,X_t^{(K)})=(X_{t,1},\ldots,X_{t,d_x})\in\mathcal X_t\subseteq\mathbb R^{d_x}$, where $d_x$ is the observation dimension. We partition the observation coordinates into $K$ groups, where $X_t^{(k)}:=X_{t,I_k}$, $I_k\subseteq[d_x]$, and $\bigsqcup_{k=1}^K I_k=[d_x]$. The number of latent coordinates expressed in group $k$ may vary with $t$. The initial state follows $Z_1\sim p_1$, where $p_1$ is supported on $\mathcal Z_1$. The first-order latent dynamics and observation model are:
\begin{align}
\label{eq:latent-dynamics}
 Z_{t,i}
 &=
 f_i\bigl(
 \mathrm{Pa}_d(Z_{t,i}),
 \epsilon_{t,i}
 \bigr),
 \quad i\in[N],
 \\
\label{eq:observation-model}
 X_t
 &=
 g_t(Z_t)\in\mathcal X_t\subseteq\mathbb R^{d_x},
\end{align}
where $f_i$ is a transition mechanism, and the noises $\{\epsilon_{t,i}\}_{t,i}$ are mutually independent with positive smooth densities $p_{\epsilon_i}$. The induced transition density $p(z_t\mid z_{t-1})$ is positive and twice continuously differentiable on its support. $\mathrm{Pa}_d(Z_{t,i})$ denotes the set of delayed parents of $Z_{t,i}$. We assume a time lag of 1 in Eq.~\ref{eq:latent-dynamics}. Instantaneous dependencies and higher-order lags are discussed in Appendix~\ref{app:extensions}.

For $t\in\{2,\ldots,T\}$, the delayed adjacency matrix $G_t^d\in\{0,1\}^{N\times N}$ is defined by $(G_t^d)_{i,j}=1$ iff $Z_{t-1,j}\in\mathrm{Pa}_d(Z_{t,i})$. Its rows therefore index coordinates at time $t$, whereas its columns index coordinates at time $t-1$. For a differentiable map $\phi$, $J_\phi$ denotes its Jacobian. In particular, $J_{g_t}(z_t)\in\mathbb R^{d_x\times N}$, whose $(r,i)$-th entry is $\partial g_{t,r}(z_t)/\partial z_{t,i}$. 
To formalize the identification problem, let $\mathcal G$ denote the DGP specified by $p_1$, $\{(f_i,p_{\epsilon_i})\}_{i=1}^N$, and $\{g_t\}_{t=1}^T$. Let $\widehat{\mathcal G}$ be a learned DGP of the same form, with latent variables $\widehat Z_t$, transition mechanisms $\widehat f_i$, noises $\widehat\epsilon_{t,i}$, observation maps $\widehat g_t$.

\begin{definition}[Coordinate subspace]
\label{def:coordinate-subspace}
For $S\subseteq[n]$, the coordinate subspace of $\mathbb R^n$ indexed by $S$ is: 
$$
\mathbb R_S^n := \{v\in\mathbb R^n:v_i=0\text{ for }i\notin S\}.
$$
\end{definition}

\begin{definition}[Support of a matrix-valued map]
\label{def:function-support}
For $F:\mathcal U\to\mathbb R^{m\times n}$, define:
$$
 \text{supp}(F)
 :=
 \bigl\{(r,i)\in[m]\times[n]:
 \exists u\in\mathcal U\ \text{s.t.}\ F(u)_{r,i}\neq0\bigr\}.
$$
\end{definition}

\begin{definition}[Matrix support space]
\label{def:matrix-support-space}
For a binary pattern $G\in\{0,1\}^{m\times n}$, define:
$$
 \mathbb R_G^{m\times n}:= \{Q\in\mathbb R^{m\times n}:G_{r,i}=0\Rightarrow Q_{r,i}=0\}.
$$
\end{definition}

\begin{definition}[Observational equivalence]
\label{def:observational-equivalence}
The models $\mathcal G$ and $\widehat{\mathcal G}$ are observationally equivalent if they induce the same distribution over observed sequences, $p_{\widehat{\mathcal G}}(X_{1:T})=p_{\mathcal G}(X_{1:T})$.
\end{definition}


\textbf{Grouped observations and structural roles.}
The observation map and delayed graph assign three complementary roles to each latent coordinate. We define the observation roles for $t\in[T]$, the incoming roles for $t\in\{2,\ldots,T\}$, and the outgoing roles for $t\in\{1,\ldots,T-1\}$ as:
\begin{align}
 \mathcal S_{t,k}^{\mathbb A}
 &:=
 \bigl\{i\in[N]:
 \exists r\in I_k\ \text{s.t.}\ (r,i)\in \text{supp}(J_{g_t})\bigr\},
 &
 \mathbb A_{t,i}^{(k)}
 &:=
 \mathbf 1\{i\in\mathcal S_{t,k}^{\mathbb A}\},
 \label{eq:A-role}\\
 \mathcal S_t^{\mathbb B}
 &:=
 \bigl\{i\in[N]:
 \exists j\in[N]\ \text{s.t.}\ (G_t^d)_{i,j}=1\bigr\},
 &
 \mathbb B_{t,i}
 &:=
 \mathbf 1\{i\in\mathcal S_t^{\mathbb B}\},
 \label{eq:B-role}\\
 \mathcal S_t^{\mathbb C}
 &:=
 \bigl\{i\in[N]:
 \exists j\in[N]\ \text{s.t.}\ (G_{t+1}^d)_{j,i}=1\bigr\},
 &
 \mathbb C_{t,i}
 &:=
 \mathbf 1\{i\in\mathcal S_t^{\mathbb C}\}.
 \label{eq:C-role}
\end{align}
Then $\mathbb A^{(k)}_{t,i}$ indicates expression in group $k$, while $\mathbb B_{t,i}$ and $\mathbb C_{t,i}$ indicate incoming and outgoing delayed influence.
For $t\in\{2,\ldots,T-1\}$, combining these roles gives $\left( \{\mathbb A_{t,i}^{(k)}\}_{k=1}^K, \mathbb B_{t,i}, \mathbb C_{t,i} \right)$. For $\boldsymbol a=(a_1,\ldots,a_K)\in\{0,1\}^K$ and
$b,c\in\{0,1\}$, the associated role class is: 
\begin{equation}
\label{eq:role-class}
 \mathcal R_t^{\boldsymbol a,b,c}
 :=
 \bigl\{i\in[N]: \mathbb A_{t,i}^{(k)}=a_k\ \forall k\in[K],\ \mathbb B_{t,i}=b,\ \mathbb C_{t,i}=c \bigr\}.
\end{equation}
Thus, the $2^{K+2} - 4$ possible signatures \footnote{We assume that a latent variable belongs to at least one group. Then we have $2^K-1$ possibilities. Together with the four possible $(\mathbb B,\mathbb C)$ patterns, this yields $4(2^K-1)=2^{K+2}-4$ joint roles.}
retain the entire group-incidence pattern together with the incoming and outgoing temporal roles. Our identification target is a single permutation $\sigma_t$ at each interior time s.t.
$\left(
\{\widehat{\mathbb A}_{t,\sigma_t(i)}^{(k)}\}_{k=1}^K,
\widehat{\mathbb B}_{t,\sigma_t(i)},
\widehat{\mathbb C}_{t,\sigma_t(i)}
\right)
=
\left(
\{\mathbb A_{t,i}^{(k)}\}_{k=1}^K,
\mathbb B_{t,i},
\mathbb C_{t,i}
\right),
\quad
\forall i\in[N]$. 


\section{Identifiability Theory} 
\label{sec:theory}

In this section, we establish identifiability of the structural roles defined above. Theorem~\ref{thm:group-support} first identifies all group-specific observation supports $\mathbb A$. Theorem~\ref{thm:outgoing-temporal-support} and Corollary~\ref{cor:incoming-temporal-support} then identify the outgoing and incoming temporal supports $\mathbb C$ and $\mathbb B$. Lemma~\ref{lem:common-support-alignment} aligns these three results, leading to joint identification of the complete signature in Theorem~\ref{thm:joint-roles}. Complete proofs are in Appendix~\ref{app:proofs}.

\subsection{Identifying grouped observation support}
\label{sec:group-support}

We first identify which latent coordinates are expressed in each observation group. The collection $\{\mathcal S_{t,k}^{\mathbb A}\}_{k=1}^K$ characterizes the complete incidence structure between latent coordinates and observation groups at time $t$, including which latents are expressed jointly or exclusively across different groups.

\begin{theorem}[Identifiability of grouped observation support]
\label{thm:group-support}
For $t\in[T]$, suppose that $\mathcal G$ and
$\widehat{\mathcal G}$ are observationally equivalent and that:
\begin{enumerate}
\renewcommand{\labelenumi}{\roman{enumi}.}
 \item
 The maps $g_t:\mathcal Z_t\to\mathcal X_t$ and $\widehat g_t:\widehat{\mathcal Z}_t\to\mathcal X_t$ are $C^1$ embeddings onto the same observation manifold. Consequently,
 $h_t=\widehat g_t^{-1}\circ g_t$ is a $C^1$ diffeomorphism.

 \item
 There exists $z_t^\star\in\mathcal Z_t$ s.t., for every $k\in[K]$, $\operatorname{span} \bigl\{(J_{g_t}(z_t^\star))_{r,:}:r\in I_k \bigr\} = \mathbb R^N_{\mathcal S_{t,k}^{\mathbb A}}.$

 \item
 The learned grouped supports are no denser overall than the true grouped supports: $\sum_{k=1}^K \left| \widehat{\mathcal S}_{t,k}^{\mathbb A} \right| \leq \sum_{k=1}^K \left| \mathcal S_{t,k}^{\mathbb A} \right|.$
\end{enumerate}

Let $\widehat z_t^\star=h_t(z_t^\star)$ and
$D_t^\star:=J_{h_t^{-1}}(\widehat z_t^\star)$. Then $(D_t^\star)_{i,v}=0, \quad \forall i\in\mathcal S_{t,k}^{\mathbb A},\;
\forall v\notin\widehat{\mathcal S}_{t,k}^{\mathbb A},\ k\in[K].$
Moreover, every permutation $\sigma_t$ s.t.
$(D_t^\star)_{i,\sigma_t(i)}\neq0 \forall i\in[N]$
aligns the group supports $\widehat{\mathcal S}_{t,k}^{\mathbb A}= \sigma_t\bigl(\mathcal S_{t,k}^{\mathbb A}\bigr), \; \forall k\in[K].$
At least one such permutation exists. 
\end{theorem}

Theorem~\ref{thm:group-support} recovers the full expression pattern under one permutation. For example, the coordinate corresponding to $Z_{t,2}$ in Figure~\ref{fig:intro_overview} is identified as being expressed in Groups 1 and 2 and absent from Group 3. 
Assumption~(i) gives a valid invertible change of latent coordinates.
Assumption~(ii) requires enough variation within each group to distinguish the latent directions it expresses. 
Conditions of this type have been widely adopted in the identifiability literature to exclude degenerate generating mechanisms and ensure that the relevant structural directions are sufficiently excited~\citep{3666122.3666707,li2025identification}.
Assumption~(iii) expresses structural minimality by allowing no more links between groups and latents than the true structure. Together, these conditions rule out indistinguishable directions and extra links between groups and latents.

\subsection{Identifying incoming and outgoing temporal support}
\label{sec:temporal-support}
Having identified the group-specific observation supports $\mathbb A$, we now turn to $\mathbb B$ and $\mathbb C$, which characterize how latent coordinates receive and transmit delayed influence across consecutive time steps.

Temporal propagation is encoded by the transition distribution between latent states. Let $q_{t+1}(z_{t+1}\mid z_t):=\log p(z_{t+1}\mid z_t)$ and define the transition cross-Hessian:
\begin{equation}
\label{eq:transition-cross-hessian}
 M_t(z_{t+1},z_t)
 :=
 \nabla_{z_{t+1}}\nabla_{z_t}^{\top}
 q_{t+1}(z_{t+1}\mid z_t).
\end{equation}
The learned cross-Hessian $\widehat M_t(\widehat z_{t+1},\widehat z_t)$ is defined analogously from the learned transition density $p_{\widehat{\mathcal G}}(\widehat z_{t+1}\mid\widehat z_t)$. Rows of $M_t$ correspond to latent coordinates at time $t+1$, whereas columns correspond to latent coordinates at time $t$. Its row and column supports therefore characterize incoming and outgoing delayed influence. We first identify the outgoing support $\mathcal S_t^{\mathbb C}$.

\begin{theorem}[Identifiability of outgoing temporal support]
\label{thm:outgoing-temporal-support}
For $t\in\{1,\ldots,T-1\}$, suppose that the true model $\mathcal G$ and learned model $\widehat{\mathcal G}$ are observationally equivalent and that:
\begin{enumerate}
\renewcommand{\labelenumi}{\roman{enumi}.}
 \item
 The transition densities $p(z_{t+1}\mid z_t), p(\widehat z_{t+1}\mid\widehat z_t)$ are positive and $C^2$, and the observation embeddings induce $C^1$ diffeomorphisms $h_t:\mathcal Z_t\to\widehat{\mathcal Z}_t$ and $h_{t+1}:\mathcal Z_{t+1}\to\widehat{\mathcal Z}_{t+1}$.

 \item
 The true and learned delayed graphs are faithful to the cross-Hessian. For every $i,j\in[N]$
 $$
 (G_{t+1}^d)_{j,i}=1 \iff \exists (z_{t+1},z_t)\in \mathcal Z_{t+1}\times\mathcal Z_t, \;\; \text{s.t.}\; [M_t(z_{t+1},z_t)]_{j,i}\neq0,
 $$
 with the analogous equivalence holding for $\widehat G_{t+1}^d$ and $\widehat M_t$.
 

 \item
 For $ \forall \;z_t^\star\in\mathcal Z_t$, $\operatorname{span}
 \bigl\{ M_t(z_{t+1},z_t^\star): z_{t+1}\in\mathcal Z_{t+1} \bigr\} = \mathbb R^{N\times N}_{G_{t+1}^d}.$

 \item
 The learned outgoing support satisfies
 $\lvert\widehat{\mathcal S}_t^{\mathbb C}\rvert
 \leq\lvert\mathcal S_t^{\mathbb C}\rvert$.
\end{enumerate}
Then, for $ \forall \;\widehat z_t\in\widehat{\mathcal Z}_t$,
$ \frac{\partial Z_{t,i}}{\partial\widehat Z_{t,v}}=0 \quad \text{for } i\in\mathcal S_t^{\mathbb C},\ v\notin\widehat{\mathcal S}_t^{\mathbb C}.$ For any $\widehat z_t^0$, let
$D_t^0:=J_{h_t^{-1}}(\widehat z_t^0)$. At least one permutation $\sigma_t$ exists s.t. $(D_t^0)_{i,\sigma_t(i)}\neq0$ for $ \forall \;i\in[N]$ satisfies $\widehat{\mathcal S}_t^{\mathbb C} = \sigma_t\bigl(\mathcal S_t^{\mathbb C}\bigr)$.
\end{theorem}

Theorem~\ref{thm:outgoing-temporal-support} identifies whether each latent has delayed influence on the next state. Thus, $\mathbb C_{t,2}=1$ in Figure~\ref{fig:intro_overview} is preserved for the corresponding learned coordinate.
Assumption~(ii) requires each delayed edge to leave a detectable cross derivative in the transition density. 
Assumption~(iii) provides enough transition variation to reveal the supported directions, and
Assumption~(iv) imposes minimality on the number of outgoing-active latent coordinates. This is weaker than constraining every delayed edge and is sufficient for identifying the outgoing structural role considered here.

The column support of $M_t(z_{t+1},z_t)$ is used in Theorem~\ref{thm:outgoing-temporal-support} to identify $\mathbb C$. 
For $M_{t-1}(z_t,z_{t-1})$, the row support reveals which coordinates receive delayed influence from $t-1$, corresponding to $\mathbb B$. 

\begin{corollary}[Identifiability of incoming temporal support]
\label{cor:incoming-temporal-support}
For $t\in\{2,\ldots,T\}$, suppose that the true model $\mathcal G$ and learned model $\widehat{\mathcal G}$ are
observationally equivalent and that:
\begin{enumerate}
\renewcommand{\labelenumi}{\roman{enumi}.}
 \item
 The smoothness and faithfulness conditions in
 Theorem~\ref{thm:outgoing-temporal-support} hold for $t-1 \to t$.
 
 \item
 For $ \forall \;z_t^\star\in\mathcal Z_t$,$ \operatorname{span} \bigl\{M_{t-1}(z_t^\star,z_{t-1}): z_{t-1}\in\mathcal Z_{t-1} \bigr\} = \mathbb R^{N\times N}_{G_t^d}$.

 \item
 The learned incoming support satisfies $\lvert\widehat{\mathcal S}_t^{\mathbb B}\rvert \leq\lvert\mathcal S_t^{\mathbb B}\rvert$.
\end{enumerate}
Then, for $ \forall \;\widehat z_t$, $ \frac{\partial Z_{t,i}}{\partial\widehat Z_{t,v}}=0 \quad \text{for } i\in\mathcal S_t^{\mathbb B},\ v\notin\widehat{\mathcal S}_t^{\mathbb B}$.
For any $\widehat z_t^0$, At least one permutation $\sigma_t$ exists, supported by the nonzero pattern of $D_t^0=J_{h_t^{-1}}(\widehat z_t^0)$ satisfies $\widehat{\mathcal S}_t^{\mathbb B} = \sigma_t\bigl(\mathcal S_t^{\mathbb B}\bigr).$
\end{corollary}

Corollary~\ref{cor:incoming-temporal-support} gives the guarantee for incoming influence. In Figure~\ref{fig:intro_overview}, it preserves $\mathbb B_{t,2}=1$, so the recovered coordinate for $Z_{t,2}$ receives influence from the past. Its assumptions play the same roles as those in Theorem~\ref{thm:outgoing-temporal-support}, now applied to the row support of the transition cross Hessian.

\subsection{Joint identification of structural roles}
\label{sec:joint-role-identification}

Theorem~\ref{thm:group-support}, Theorem~\ref{thm:outgoing-temporal-support}, and Corollary~\ref{cor:incoming-temporal-support} identify $\mathbb A$, $\mathbb C$, and $\mathbb B$ separately up to permutation. We next show that these structures can be aligned by one permutation at the same time point.

\begin{lemma}[Common support alignment]
\label{lem:common-support-alignment}
For $t\in\{2,\ldots,T-1\}$, suppose that the assumptions of
Theorem~\ref{thm:group-support}, Theorem~\ref{thm:outgoing-temporal-support}, and Corollary~\ref{cor:incoming-temporal-support} hold. Let $z_t^\star\in\mathcal Z_t$ be the point in assumption~(ii) of Theorem~\ref{thm:group-support}, and define $\widehat z_t^\star:=h_t(z_t^\star)$ and $D_t^\star:=J_{h_t^{-1}}(\widehat z_t^\star)$. There exists at least one permutation $\sigma_t:[N]\to[N]$ s.t. $(D_t^\star)_{i,\sigma_t(i)}\neq 0 \;$ for $\; \forall \;i\in[N], \; \forall \; k\in[K]$ simultaneously satisfies:
\begin{align}
 \widehat{\mathcal S}_{t,k}^{\mathbb A} = \sigma_t\bigl(\mathcal S_{t,k}^{\mathbb A}\bigr),\;\;
 \widehat{\mathcal S}_t^{\mathbb B} = \sigma_t\bigl(\mathcal S_t^{\mathbb B}\bigr), \;\;
 \widehat{\mathcal S}_t^{\mathbb C} = \sigma_t\bigl(\mathcal S_t^{\mathbb C}\bigr).
\end{align}
\end{lemma}

Lemma~\ref{lem:common-support-alignment} resolves the permutation ambiguity in Figure~\ref{fig:intro_overview} (c). Intuitively, the three preceding results all constrain the row support of the same time-$t$ reparameterization Jacobian $J_{h_t^{-1}}$. Evaluating them at a common point therefore yields a single permutation that aligns all structures. Hence, the group, incoming, and outgoing roles attached to one coordinate correspond to the same true latent.

\begin{theorem}[Joint identifiability of temporal latent roles]
\label{thm:joint-roles}
Under the conditions of Lemma~\ref{lem:common-support-alignment}, there exists a permutation $\sigma_t:[N]\to[N]$ s.t., for $ \forall \;i\in[N]$,
\begin{equation}
\left( \left\{ \widehat{\mathbb A}_{t,\sigma_t(i)}^{(k)} \right\}_{k=1}^K, \widehat{\mathbb B}_{t,\sigma_t(i)}, \widehat{\mathbb C}_{t,\sigma_t(i)} \right) = \left( \left\{ \mathbb A_{t,i}^{(k)} \right\}_{k=1}^K, \mathbb B_{t,i}, \mathbb C_{t,i} \right).
\label{eq:joint-indicators}
\end{equation}
Consequently, for every $\boldsymbol a\in\{0,1\}^K$ and $b,c\in\{0,1\}$, $\widehat{\mathcal R}_t^{\boldsymbol a,b,c} = \sigma_t\bigl(\mathcal R_t^{\boldsymbol a,b,c}\bigr)$.
\end{theorem}

\paragraph{Discussion.}
Theorem~\ref{thm:joint-roles} establishes a common correspondence between group expression and temporal influence. Lemma~\ref{lem:common-support-alignment} obtains this correspondence from one nonsingular Jacobian $D_t^\star$, yielding joint support identification without a path-connectedness assumption.~\citep{3692070.3692351, lachapelle2023additive} use continuity and connectedness, together with local Jacobian structure, to keep a coordinate or block permutation fixed across latent states. Our result aligns supports defined over the entire latent domain while allowing ambiguity among coordinates with the same structural role. This suffices to identify all admissible role classes at time $t$. Appendix~\ref{app:common-vs-global} gives a geometric explanation and compares this guarantee with related temporal identification results.


\section{\textsc{RoleTrace}: Estimation of Temporal Latent Roles}
\label{sec:estimation}

Guided by the identifiability results in Section~\ref{sec:theory}, we propose \textsc{RoleTrace}, a practical framework for estimating latent structural roles from finite data. \textsc{RoleTrace} 
first learn a latent representation and its temporal transition density, estimate the three structural roles from the corresponding derivatives, and finally refine a common latent coordinate system in which the three roles are jointly sparse. 

We use an encoder--decoder pair to estimate the latent representation underlying the grouped observations $X_t=(X_t^{(1)},\ldots,X_t^{(K)})$. Let $e_{\omega}$ denote the encoder and $\widetilde g_t$ the decoder. 
We define $\widetilde Z_t=e_{\omega}(X_t)\in\mathbb R^N$ and reconstruct $\widetilde X_t=\widetilde g_t(\widetilde Z_t)$. 
Given $\widetilde Z_{1:T}$, we model the transition with a differentiable conditional density $\widetilde p_{\eta}(\widetilde z_{t+1}\mid\widetilde z_t)$ whose log density has mixed second derivatives:  
\begin{equation}
\begin{aligned}
\mathcal L_{\mathrm{rec}}
&=
\frac{1}{T}\sum_{t=1}^{T}
\ell_x\!\left(X_t,\widetilde g_t(\widetilde Z_t)\right),
\quad
\mathcal L_{\mathrm{trans}}
&=
-\frac{1}{T-1}\sum_{t=1}^{T-1}
\log \widetilde p_{\eta}
\!\left(\widetilde Z_{t+1}\mid\widetilde Z_t\right).
\end{aligned}
\label{eq:est_objectives}
\end{equation}
Here, $\ell_x$ denotes the reconstruction loss. We first learn the latent representation and keep it fixed during temporal density estimation, which preserves the geometry of the learned latent space.

\paragraph{Structural Role Estimation.}
We estimate the three structural roles from the derivative objects appearing in Section~\ref{sec:theory}. Define the decoder Jacobian $\widetilde J_t(\widetilde z_t):=J_{\widetilde g_t}(\widetilde z_t)$ and let $ \widetilde q_{t+1}(\widetilde z_{t+1}\mid\widetilde z_t) := \log\widetilde p_{\eta}(\widetilde z_{t+1}\mid\widetilde z_t), $ with transition cross-Hessian $ \widetilde M_t(\widetilde z_{t+1},\widetilde z_t) := \nabla_{\widetilde z_{t+1}} \nabla_{\widetilde z_t}^{\top} \widetilde q_{t+1}(\widetilde z_{t+1}\mid\widetilde z_t)$. Its rows correspond to coordinates at time $t+1$, while its columns correspond to coordinates at time $t$.

For each role, we extract the derivative slice associated with latent coordinate $i$:
\begin{equation}
\begin{aligned}
v^{\mathbb A}_{t,k,i} 
&:= [\widetilde J_t(\widetilde Z_t)]_{I_k,i},
&\quad
v^{\mathbb B}_{t,i}
&:= [\widetilde M_{t-1}(\widetilde Z_t,\widetilde Z_{t-1})]_{i,:},
&\quad
v^{\mathbb C}_{t,i}
&:= [\widetilde M_t(\widetilde Z_{t+1},\widetilde Z_t)]_{:,i},
\\
d^{\mathbb A}_{k} 
&:= |I_k|,
&\quad
d^{\mathbb B}
&:= N,
&\quad
d^{\mathbb C}
&:= N.
\end{aligned}
\label{eq:role_derivative_slices}
\end{equation}
We then use a common root-mean-square score:
\begin{equation}
s^{\mathbb X} = \left[ \mathbb E_{\mathrm{emp}} \left( \frac{1}{d^{\mathbb X}} \left\| v^{\mathbb X} \right\|_2^2 \right) \right]^{1/2},
\quad \mathbb X\in\{\mathbb A,\mathbb B,\mathbb C\},
\label{eq:role_score}
\end{equation}
with the indices inherited from Eq.~\ref{eq:role_derivative_slices}. Thus, $s^{\mathbb A}_{t,k,i}$ measures how strongly latent coordinate $i$ is expressed in group $k$, $s^{\mathbb B}_{t,i}$ and $s^{\mathbb C}_{t,i}$ measures its incoming and outgoing temporal influence. These scores are continuous empirical surrogates for the support conditions defining $\mathcal S^{\mathbb A}_{t,k}$, $\mathcal S^{\mathbb B}_t$, and $\mathcal S^{\mathbb C}_t$. After structural refinement, we threshold the three score types to obtain $\widehat{\mathbb A}^{(k)}_{t,i}$, $\widehat{\mathbb B}_{t,i}$, and $\widehat{\mathbb C}_{t,i}$.

\paragraph{Common Coordinate Structural Refinement.}
The derivative scores above depend on the coordinates selected by the initial representation learner. Motivated by Theorem~\ref{thm:joint-roles}, we search for a single refined coordinate system at each time point that makes the three structural views jointly sparse.

For $t\in\{2,\ldots,T-1\}$, let $Q_t\in\mathbb R^{N\times N}$ be an orthogonal transformation, $Q_t^{\top}Q_t=I$, and define the latent coordinates by $\widehat Z_t = Q_t^{\top}\widetilde Z_t.$ Equivalently, $\widetilde Z_t=Q_t\widehat Z_t$. The observation decoder in the coordinates is therefore $\widehat g_t(\widehat z_t)=\widetilde g_t(Q_t\widehat z_t)$, which gives $ J_{\widehat g_t}(\widehat z_t) = \widetilde J_t(\widetilde z_t)Q_t.$ For the transitions $t-1\rightarrow t$ and $t\rightarrow t+1$, the cross-Hessians transform on the time $t$ coordinate as $ \widehat M_{t-1} = Q_t^{\top}\widetilde M_{t-1}, \; \widehat M_t = \widetilde M_t Q_t.$
Hence, the same $Q_t$ simultaneously controls the group-incidence columns of the observation Jacobian, the incoming rows of $\widetilde M_{t-1}$, and the outgoing columns of $\widetilde M_t$.
Let $s^{\mathbb A}_{t,k,i}(Q_t)$, $s^{\mathbb B}_{t,i}(Q_t)$, and $s^{\mathbb C}_{t,i}(Q_t)$ denote the resulting scores. We use the sparsity objective:
\begin{align}
 \Omega_t(Q_t) = \lambda_{\mathbb A} \sum_{k=1}^{K}\sum_{i=1}^{N} s^{\mathbb A}_{t,k,i}(Q_t) 
 + \lambda_{\mathbb B} \sum_{i=1}^{N} s^{\mathbb B}_{t,i}(Q_t)
 + \lambda_{\mathbb C} \sum_{i=1}^{N} s^{\mathbb C}_{t,i}(Q_t),
 \label{eq:joint_structural_objective}
\end{align}
This finite-sample sparsity objective is a practical surrogate for the minimality conditions in Section~\ref{sec:theory}. We minimize Eq.~\ref{eq:joint_structural_objective} using deterministic Jacobi coordinate rotations. Let $G_{ij}(\theta)$ denote the identity matrix with its $(i,j)$ block replaced by a rotation of angle $\theta$. For each pair $(i,j)$, we update
$ \theta^{\star} = \arg\min_{\theta} \Omega_t \left( Q_tG_{ij}(\theta) \right),\; Q_t \leftarrow Q_tG_{ij}(\theta^{\star}),$
and sweep over coordinate pairs until the structural objective no longer decreases appreciably. 
Because the estimated transition density may change after the latent coordinates are refined, we perform one additional transition refit: after obtaining the initial $Q_t$, we refit the conditional density in the coordinates, recompute the cross-Hessians, and run a final Jacobi sweep. The resulting representation yields the joint estimated signature.


\section{Experiments}
\label{sec:exp}
We conduct experiments in both synthetic and real-world settings, guided by four research questions:
\textbf{RQ1 (Recovery).} Can \textsc{RoleTrace} recover $\{\mathbb A^{(k)}\}_{k=1}^{K}$, $\mathbb B$ and $\mathbb C$?  
\textbf{RQ2 (Alignment).} Does common alignment enable more consistent joint recovery than aligning the three structural roles separately?
\textbf{RQ3 (Sensitivity to Identifiability Conditions).} How does recovery change as the key conditions underlying our identifiability results are weakened?
\textbf{RQ4 (Applicability).} Does \textsc{RoleTrace} capture meaningful information flow in real-world scenario?


\subsection{Synthetic Experiments}
\label{sec:synthetic}
\begin{figure}[t]
    \centering


    \begin{subfigure}[t]{0.49\textwidth}
    \vspace{0pt}
    \vspace{0.08in}
        \centering
        \includegraphics[width=\linewidth]{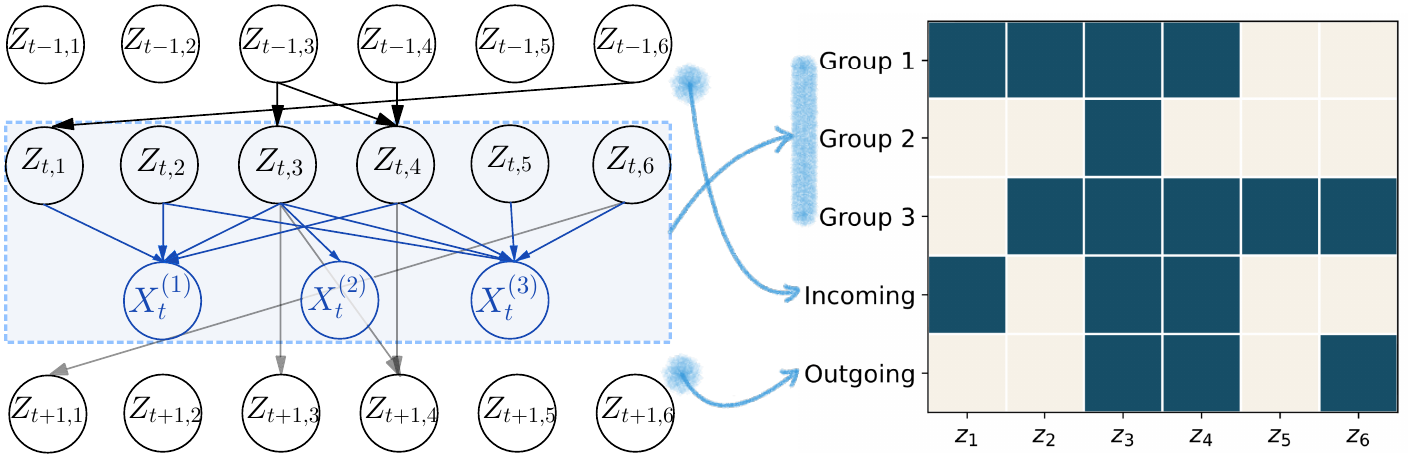}
        
        \vspace{0.04in}
        \caption{Example of DGP and ground truth.}
        \label{fig:exp-rq1-dgp}
    \end{subfigure}
    \hfill
    \begin{minipage}[t]{0.49\textwidth}
        \vspace{0pt}
        \centering

        \begin{subfigure}[t]{0.48\linewidth}
            \centering
            \includegraphics[width=\linewidth]{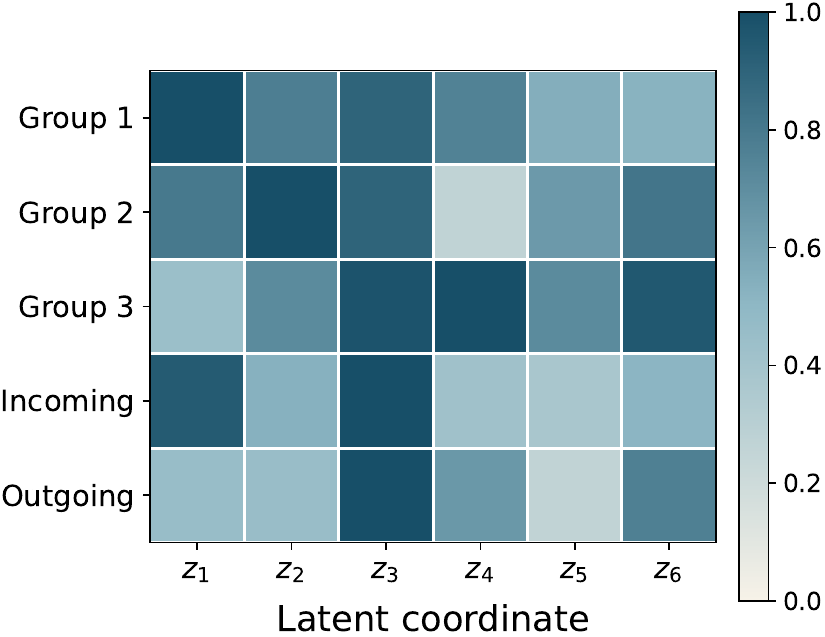}
            \caption{No structural}
            \label{fig:rq1-heatmap-no-structural}
        \end{subfigure}
        \hfill
        \begin{subfigure}[t]{0.48\linewidth}
            \centering
            \includegraphics[width=\linewidth]{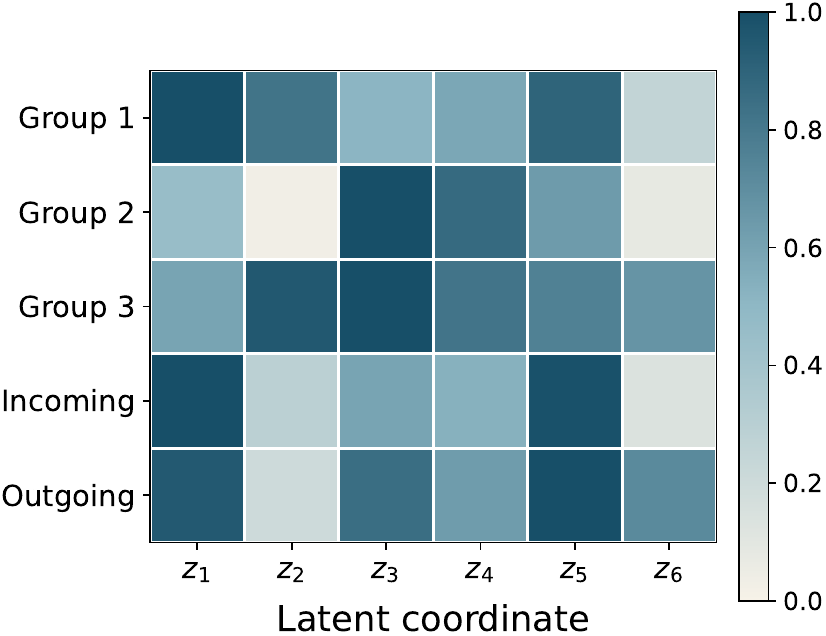}
            \caption{Temporal only}
            \label{fig:rq1-heatmap-temporal-only}
        \end{subfigure}

    \end{minipage}

    \vspace{0.5em}


    \begin{subfigure}[t]{0.235\textwidth}
        \centering
        \includegraphics[width=\linewidth]{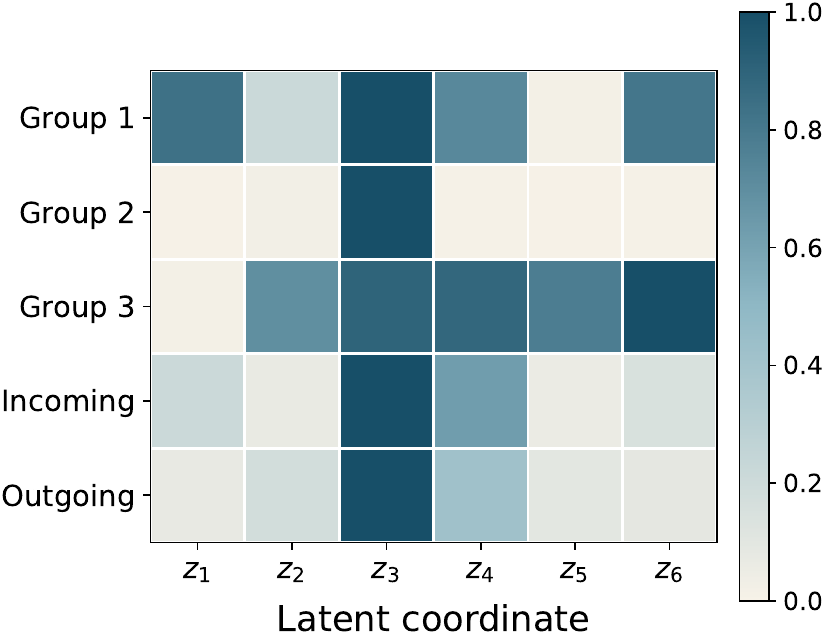}
        \caption{No refit}
        \label{fig:rq1-heatmap-no-refit}
    \end{subfigure}
    \hfill
    \begin{subfigure}[t]{0.235\textwidth}
        \centering
        \includegraphics[width=\linewidth]{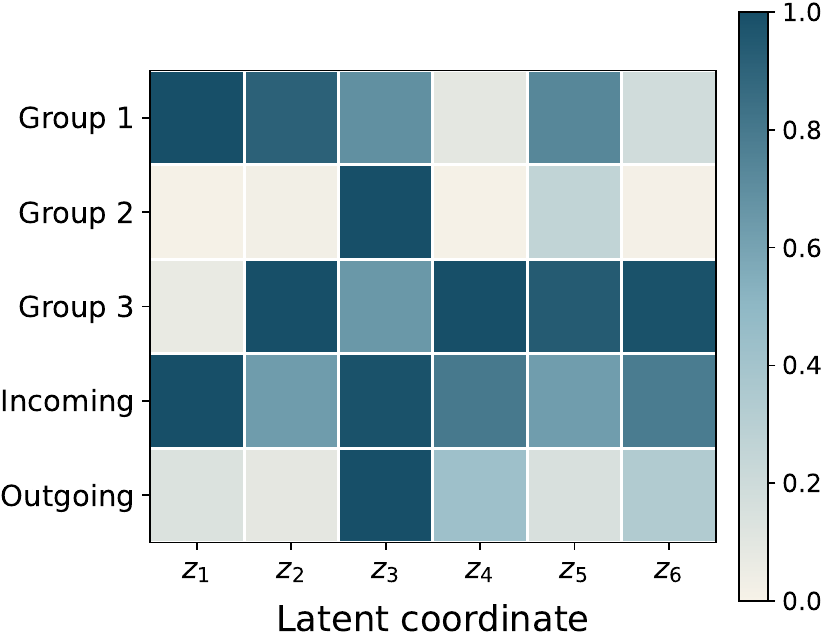}
        \caption{InterLatent}
        \label{fig:rq1-heatmap-InterLatent}
    \end{subfigure}
    \hfill
    \begin{subfigure}[t]{0.235\textwidth}
        \centering
        \includegraphics[width=\linewidth]{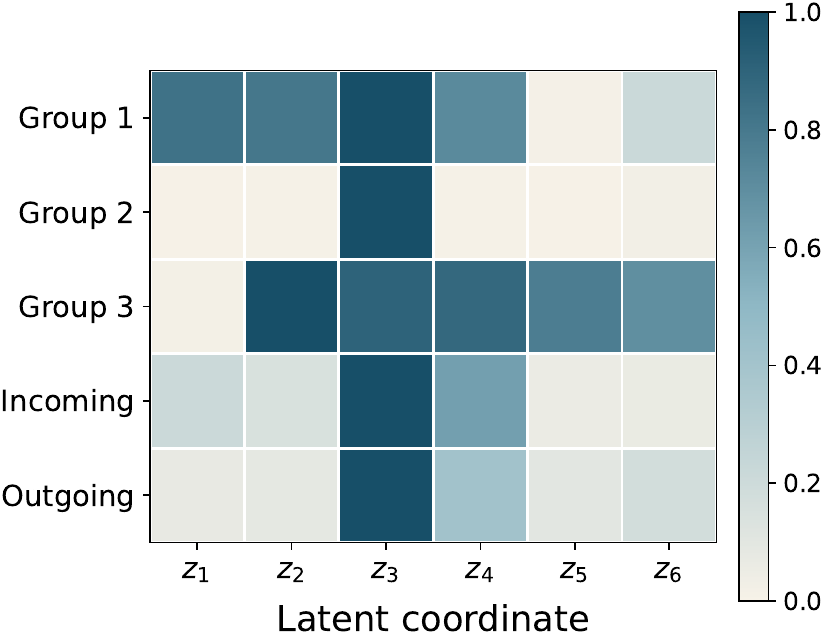}
        \caption{ThoughtComm}
        \label{fig:rq1-heatmap-thoughtcomm}
    \end{subfigure}
    \hfill
    \begin{subfigure}[t]{0.235\textwidth}
        \centering
        \includegraphics[width=\linewidth]{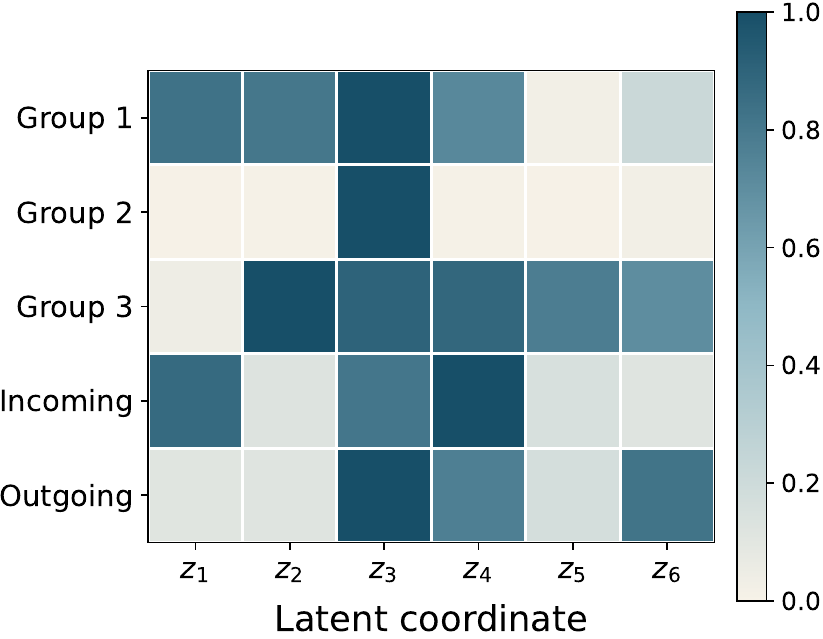}
        \caption{RoleTrace}
        \label{fig:rq1-heatmap-ours}
    \end{subfigure}

    \caption{Heatmaps of recovered structural signatures at $n=3200$.}
    \label{fig:rq1-heatmaps}
\end{figure}

\paragraph{Setup.}
We evaluate structural role recovery across randomly generated mechanisms. The RQ1 datasets are generated by a random nonlinear latent process with Gaussian noise and a nonlinear grouped observation map. Both structural supports and numerical parameters are sampled for each mechanism; Figure~\ref{fig:exp-rq1-dgp} illustrates one example. We use 16 mechanisms with $N=6$, $K=3$, $T=5$, and six observed dimensions per group. RQ2 and RQ3 use controlled mechanisms with $N=4$.
Details of DGP, training, and evaluation metrics are provided in Appendix~\ref{app:synthetic-setup}.



\textbf{Baselines and evaluation.}
We compare two adapted methods and three ablations. \textsc{ThoughtComm} uses the observation structure $\mathbb A$ for coordinate refinement~\citep{zheng2025thought}. \textsc{InterLatent} jointly trains the representation, temporal density, and structural penalties~\citep{li2025identification2}. \textsc{No Structural} omits coordinate refinement; \textsc{Temporal-only} uses only $\mathbb B$ and $\mathbb C$; and \textsc{No-refit} omits the transition refit and final refinement. Methods share data and matched component initializations, with a common derivative readout for all three roles. We measure structural recovery under a common evaluation permutation. Role F1 scores assess $\mathbb A$, $\mathbb B$, and $\mathbb C$. Joint accuracy counts coordinates with a correct signature, while Hamming accuracy counts correct signature entries.

\begin{figure}[htbp]
    \centering

    \begin{minipage}[c]{0.55\textwidth}
        \centering
        \begin{subfigure}[c]{\linewidth}
            \centering
            \includegraphics[width=\linewidth]{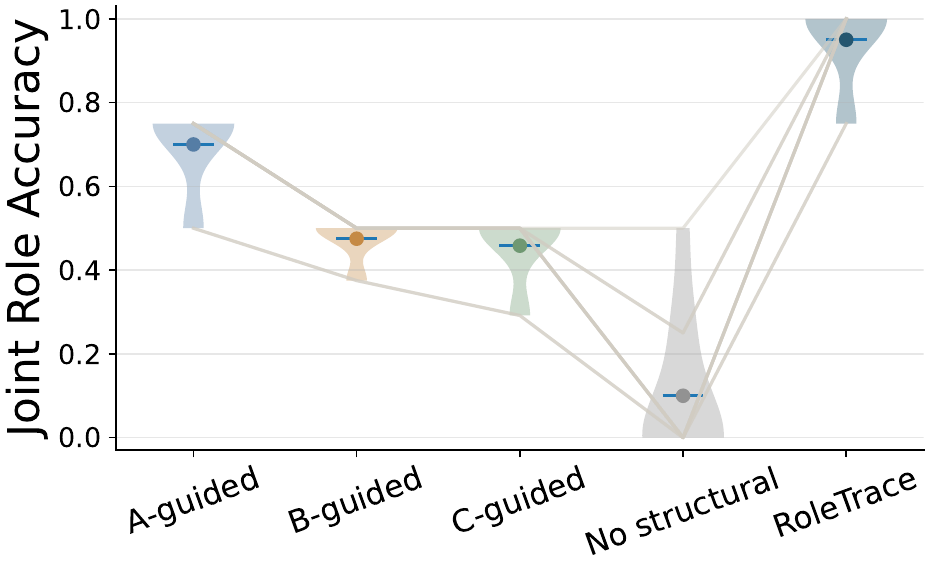}
            \caption{Joint role recovery.}
            \label{fig:rq2_joint_violin}
        \end{subfigure}
    \end{minipage}
    \hfill
    \begin{minipage}[c]{0.4\textwidth}
        \centering

        \begin{subfigure}[t]{0.36\linewidth}
            \centering
            \includegraphics[width=\linewidth]
            {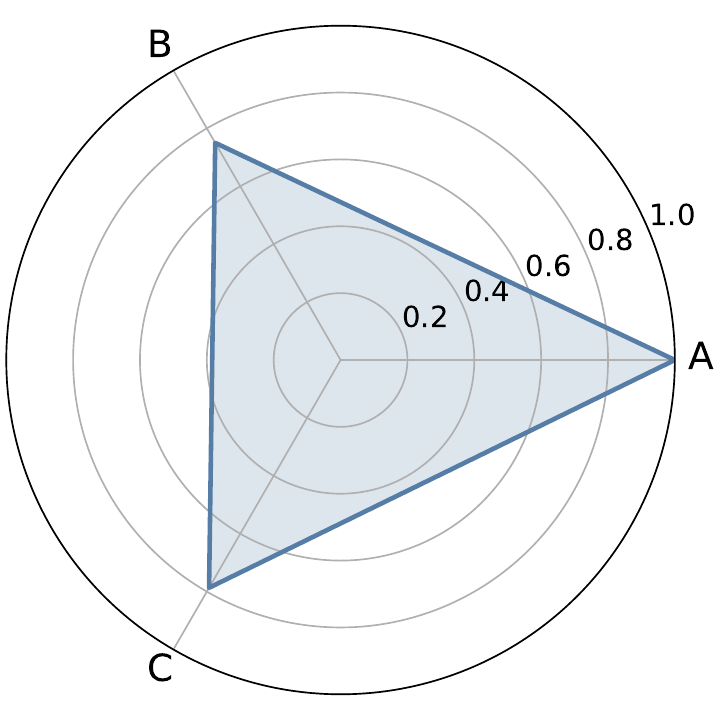}
            \caption{$\mathbb{A}$-guided.}
            \label{fig:rq2_radar_a}
        \end{subfigure}
        \hfill
        \begin{subfigure}[t]{0.36\linewidth}
            \centering
            \includegraphics[width=\linewidth]
            {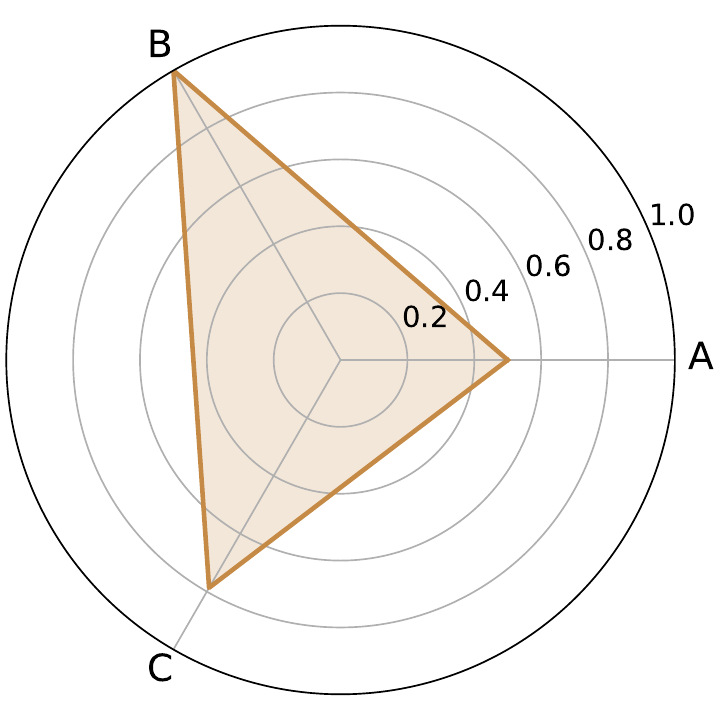}
            \caption{$\mathbb{B}$-guided.}
            \label{fig:rq2_radar_b}
        \end{subfigure}


        \begin{subfigure}[t]{0.36\linewidth}
            \centering
            \includegraphics[width=\linewidth]
            {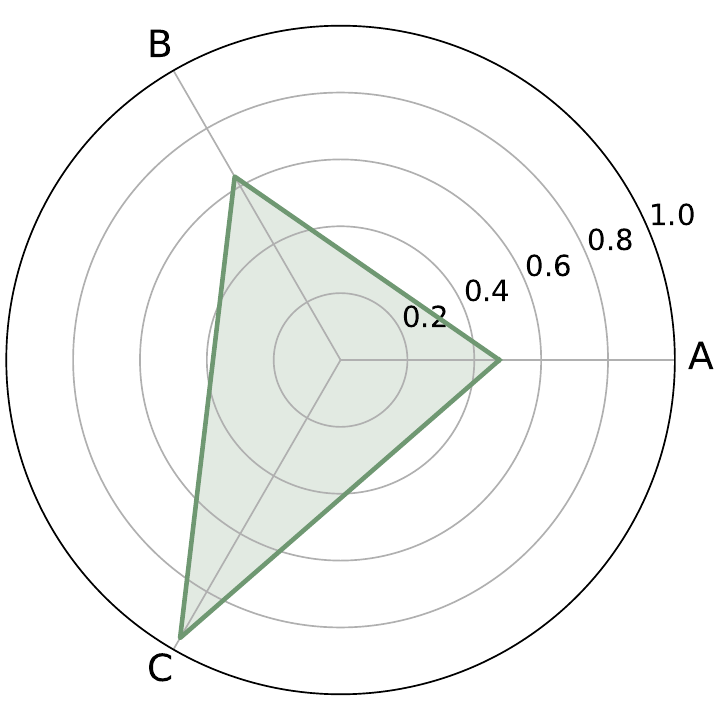}
            \caption{$\mathbb{C}$-guided.}
            \label{fig:rq2_radar_c}
        \end{subfigure}
        \hfill
        \begin{subfigure}[t]{0.36\linewidth}
            \centering
            \includegraphics[width=\linewidth]
            {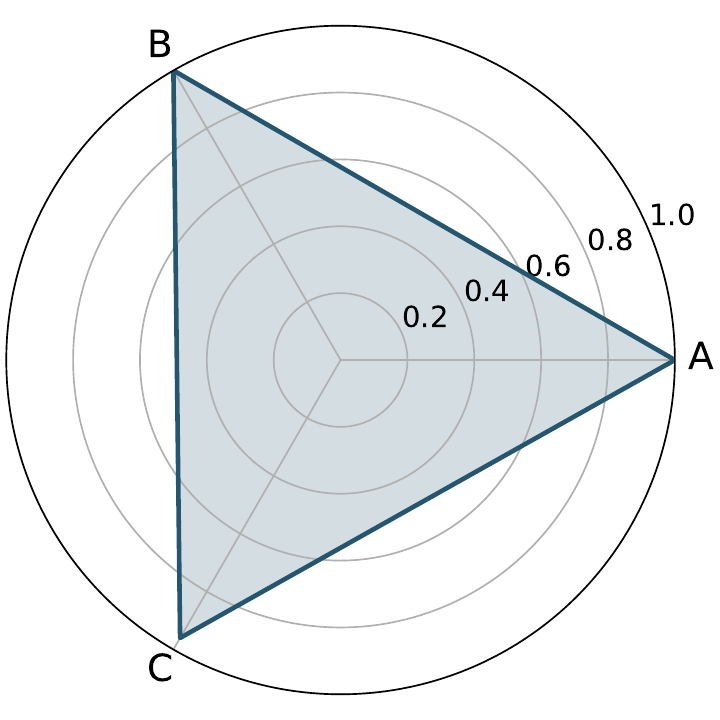}
            \caption{RoleTrace.}
            \label{fig:rq2_radar_ours}
        \end{subfigure}

    \end{minipage}

    \caption{
        Structural role identification under different permutations.
    }
    \label{fig:rq2}
\end{figure}

\begin{figure}[htbp]
\centering
\begin{subfigure}{0.3\textwidth}
    \centering
    \includegraphics[width=0.8\linewidth]{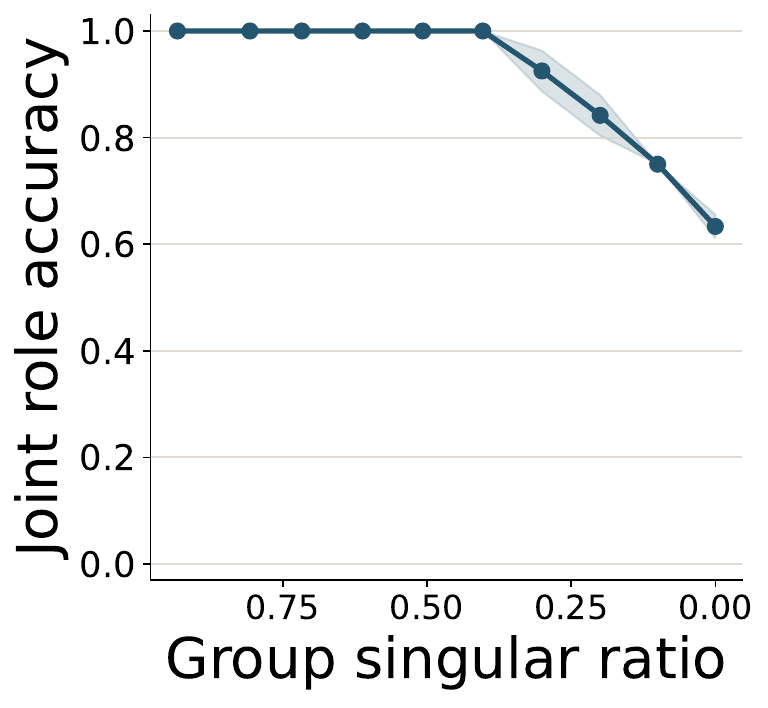}
    \caption{Group conditioning.}
\end{subfigure}
\begin{subfigure}{0.3\textwidth}
    \centering
    \includegraphics[width=0.8\linewidth]{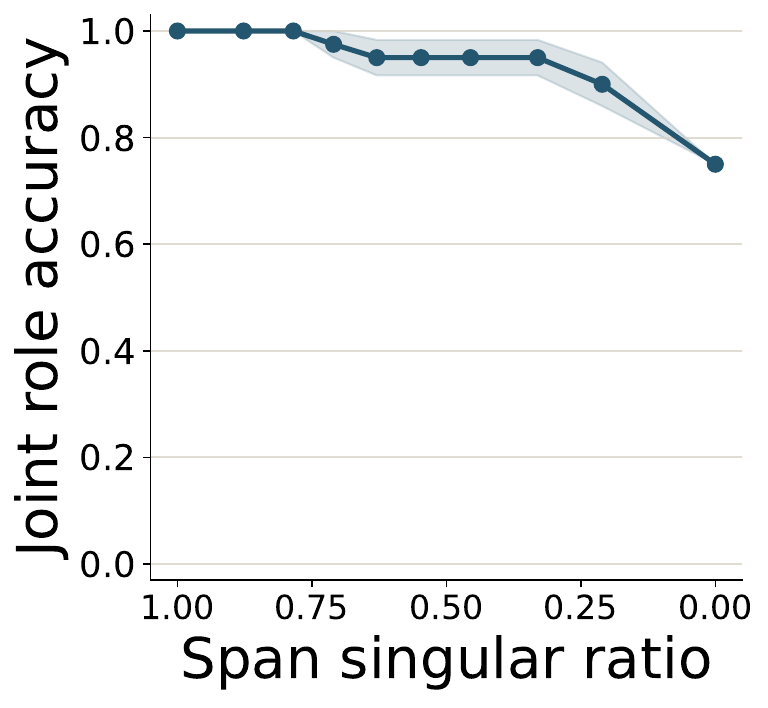}
    \caption{Temporal span.}
\end{subfigure}
\begin{subfigure}{0.3\textwidth}
    \centering
    \includegraphics[width=0.8\linewidth]{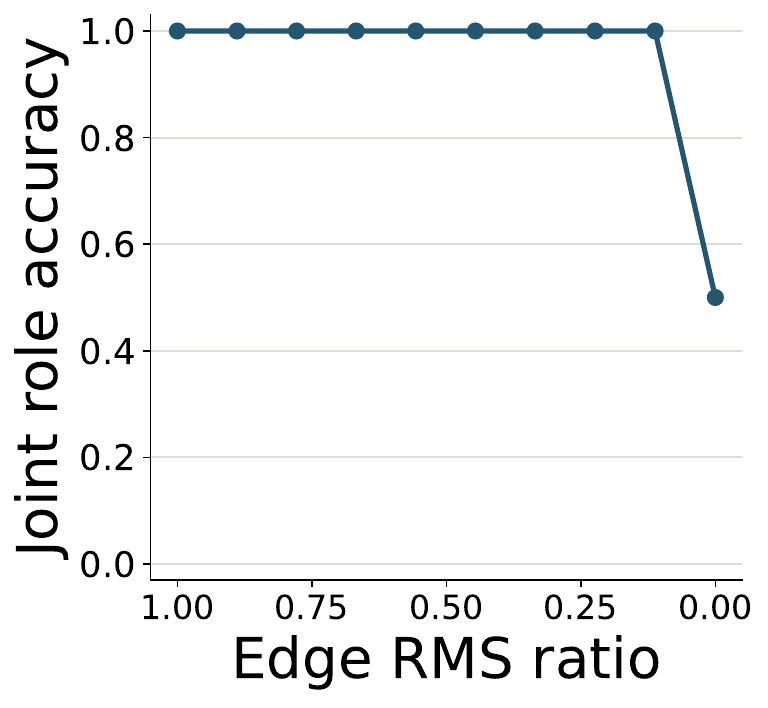}
    \caption{Edge faithfulness.}
\end{subfigure}
\caption{Sensitivity to structural conditions using true derivatives.}
\label{fig:rq3}
\end{figure}

In Table~\ref{tab:rq1-mainv2}, \textsc{RoleTrace} achieves the highest mean joint recovery among the compared methods. At $n=3200$, its Joint accuracy is $0.854$, compared with $0.708$ for both \textsc{ThoughtComm} and \textsc{No-refit}. It also reaches $0.973$ F1 for both temporal roles, while \textsc{ThoughtComm} has slightly higher $\mathbb A$ F1. The improvement over \textsc{No-refit} supports refitting the transition after coordinate refinement. Figure~\ref{fig:rq1-heatmaps} illustrates the recovered signatures in detail for the example mechanism \textbf{(Answering RQ1)}.
Joint alignment reduces the correspondence ambiguity left by individual roles. In Figure~\ref{fig:rq2}, we hold the recovered scores fixed and select evaluation permutations using $\mathbb A$, $\mathbb B$, $\mathbb C$, or all three jointly. Averaging over equally optimal permutations, the respective individual alignments give Joint accuracies of $0.700$, $0.475$, and $0.458$, compared with $0.950$ for joint alignment. Thus, accurate matching of one role can leave the complete signature unresolved \textbf{(Answering RQ2)}.
Figure~\ref{fig:rq3} examines recovery as the identifiability conditions weaken. Using true derivatives under orthogonal coordinate mixing, we vary group conditioning, temporal span, and edge faithfulness separately. Mean Joint accuracy decreases from $1$ to $0.633$, $0.750$, and $0.500$ at the respective endpoints. The first two paths show losses before the endpoint. In the cancellation experiment, the declared target remains fixed, and recovery stays perfect until the selected edge has zero net influence. These results characterize sensitivity along the tested degradation paths \textbf{(Answering RQ3)}.

\begin{table}[t]
    \centering
    \caption{Structural recovery at $n=3200$ (mean $\pm$ SE over 16 paired mechanisms).}
    \label{tab:rq1-mainv2}

    \begingroup
    \small
    \setlength{\tabcolsep}{2.6pt}
    \renewcommand{\arraystretch}{1.08}

    \begin{tabular*}{\linewidth}{
        @{\extracolsep{\fill}}lccccc@{}
    }
        \toprule
        Method
        & $\mathbb A$ F1
        & $\mathbb B$ F1
        & $\mathbb C$ F1
        & Joint Acc.
        & Hamming \\
        \midrule

        No structural
        & $0.796 \pm 0.027$
        & $0.823 \pm 0.034$
        & $0.795 \pm 0.043$
        & $0.281 \pm 0.036$
        & $0.769 \pm 0.015$ \\

        Temporal-only
        & $0.850 \pm 0.025$
        & $0.783 \pm 0.033$
        & $0.899 \pm 0.040$
        & $0.469 \pm 0.051$
        & $0.819 \pm 0.021$ \\

        No-refit
        & $0.950 \pm 0.020$
        & $0.880 \pm 0.037$
        & $0.886 \pm 0.023$
        & $0.708 \pm 0.058$
        & $0.919 \pm 0.020$ \\

        InterLatent
        & $0.847 \pm 0.017$
        & $0.717 \pm 0.048$
        & $0.833 \pm 0.032$
        & $0.323 \pm 0.042$
        & $0.790 \pm 0.016$ \\

        ThoughtComm
        & $\mathbf{0.964 \pm 0.013}$
        & $0.859 \pm 0.039$
        & $0.848 \pm 0.027$
        & $0.708 \pm 0.058$
        & $0.919 \pm 0.017$ \\

        RoleTrace
        & $0.957 \pm 0.017$
        & $\mathbf{0.973 \pm 0.014}$
        & $\mathbf{0.973 \pm 0.014}$
        & $\mathbf{0.854 \pm 0.050}$
        & $\mathbf{0.954 \pm 0.017}$ \\

        \bottomrule
    \end{tabular*}

    \endgroup
\end{table}




\subsection{Real-world Case: Multi-party Meetings}

\paragraph{Dataset and setup.}
We consider the AMI Meeting Corpus, a multimodal corpus containing 100 hours of recorded meetings with manual transcripts and linguistic annotations~\citep{10.1007/11677482_3}. 
We focus on scenario-based meetings in which four participants collaborate on a shared design project over multiple project phases. Each participant is assigned one fixed functional role: Project Manager (PM), Industrial Designer (ID), User Interface Designer (UI), and Marketing Expert (ME).  The four participant roles provide natural observation groups for an underlying conversational state that evolves throughout the meeting, which closely matches our grouped temporal setting. Let $\mathcal K_{\mathrm{AMI}} :=\{\mathrm{PM},\mathrm{ID},\mathrm{UI},\mathrm{ME}\}$, with $K=4$. At each time step $t$, we write $ X_t = \left( X_t^{(\mathrm{PM})}, X_t^{(\mathrm{ID})}, X_t^{(\mathrm{UI})}, X_t^{(\mathrm{ME})} \right) \in\mathbb R^{4\times d}$. We let $Z_t\in\mathbb R^N$ denote the latent conversational state underlying these grouped observations. After requiring the meetings to have the corresponding \textsc{RoleTrace} representations and downstream labels, our evaluation contains 118 meetings from 30 teams. We partition each meeting into 15-second windows and evaluate next-speaker prediction. Further preprocessing and evaluation details are provided in Appendix~\ref{app:ami_details}. 

We evaluate whether the identified outgoing role can retain information useful for next-speaker prediction under limited latent budgets. All methods select $k$ coordinates from the same representation and use the same downstream classifier.
\textsc{Random} samples coordinates uniformly. \textsc{Variance} keeps those with the largest variation on fit data. \textsc{Activity} ranks coordinates by their sensitivity to current speech activity. \textsc{Content} uses the corresponding temporal readout for linguistic content, and \textsc{Density} uses sensitivity from the learned transition density. Our Role $\mathbb C$ selector prioritizes coordinates whose current variation is more relevant to subsequent interaction. 
Table~\ref{tab:ami-c-comparison} shows that Role $\mathbb C$ performs best at the tighter budgets $k=4,8,12$, consistently outperforming random, variance, and activity selection. At $k=16$, it remains comparable to the strongest density readout, while at $k=24$ the three temporal selectors retain the same coordinates and therefore achieve the same accuracy of $51.475\pm0.119\%$. These results show that the outgoing structural role learned by \textsc{RoleTrace} can benefit the downstream next-speaker task by preserving more predictive information when the latent representation is compressed \textbf{(Answering RQ4)}.

\begin{table}[t]
    \centering
    \caption{
        AMI next-speaker accuracy
        (\%, mean $\pm$ standard deviation over ten seeds) at matched latent budgets.
        Additional comparisons are
        provided in Appendix~\ref{app:ami_details}.
    }
    \label{tab:ami-c-comparison}

    \begingroup
    \small
    \setlength{\tabcolsep}{2.6pt}
    \renewcommand{\arraystretch}{1.08}

    \begin{tabular*}{\linewidth}{
        @{\extracolsep{\fill}}lccccc@{}
    }
        \toprule
        Selection rule
        & $k=4$
        & $k=8$
        & $k=12$
        & $k=16$
        & $k=24$ \\
        \midrule

        Random
        & $42.822 \pm 0.588$
        & $47.528 \pm 0.259$
        & $49.341 \pm 0.154$
        & $50.190 \pm 0.105$
        & $51.108 \pm 0.088$ \\

        Variance
        & $42.442 \pm 0.748$
        & $47.139 \pm 0.584$
        & $49.222 \pm 0.338$
        & $50.118 \pm 0.310$
        & $51.077 \pm 0.126$ \\

        Activity
        & $44.774 \pm 0.936$
        & $48.031 \pm 0.522$
        & $49.686 \pm 0.450$
        & $50.438 \pm 0.293$
        & $51.197 \pm 0.289$ \\

        Content
        & $44.261 \pm 0.875$
        & $49.016 \pm 0.566$
        & $50.299 \pm 0.256$
        & $50.841 \pm 0.289$
        & $\mathbf{51.475 \pm 0.119}$ \\

        Density
        & $45.155 \pm 1.397$
        & $49.430 \pm 0.493$
        & $50.503 \pm 0.266$
        & $\mathbf{50.954 \pm 0.218}$
        & $\mathbf{51.475 \pm 0.119}$ \\


        Role $\mathbb C$
        & $\mathbf{45.854 \pm 1.413}$
        & $\mathbf{49.470 \pm 0.384}$
        & $\mathbf{50.546 \pm 0.317}$
        & $50.906 \pm 0.261$
        & $\mathbf{51.475 \pm 0.119}$ \\

        \bottomrule
    \end{tabular*}

    \endgroup
\end{table}

\section{Conclusion}
\label{sec:conclusion}
We study the joint identification of latent structural roles from grouped sequential observations. Under suitable structural and variability conditions, we show that group expression $\mathbb A$, incoming influence $\mathbb B$, and outgoing influence $\mathbb C$ are jointly identifiable under a common permutation $\sigma_t$, recovering their correspondence at each latent coordinate. Guided by these results, we develop \textsc{RoleTrace} to estimate the roles from finite data in a shared latent coordinate system. Synthetic experiments support structural recovery and highlight the importance of common alignment, while the AMI results suggest that our approach is beneficial for downstream tasks.
Together, our theory and framework provide a foundation for understanding where hidden factors are expressed and how their influence propagates through evolving systems. 
Although we have not explored a broader range of real-world applications empirically, the proposed formulation applies to a broader class of grouped sequential systems than those evaluated in this work. We hope this work sheds light on the latent organization underlying sequential observations and opens new directions for understanding complex evolving systems through identifiable representations.



\bibliography{iclr2027_conference}
\bibliographystyle{iclr2027_conference}

\appendix
\addcontentsline{toc}{section}
{Appendix}
\section*{Table of Contents}

\begingroup
\hypersetup{hidelinks}
\small

\makeatletter

\newcommand{\apptocsection}[3]{%
  \@dottedtocline{1}{0em}{2.1em}%
    {\hyperref[#1]{\numberline{\textbf{#2}}#3}}%
    {\pageref*{#1}}%
}

\newcommand{\apptocsubsection}[3]{%
  \@dottedtocline{2}{1.45em}{3.0em}%
    {\hyperref[#1]{\numberline{#2}#3}}%
    {\pageref*{#1}}%
}

\apptocsection{app:proofs}{A}{Proofs}
\apptocsubsection{app:proof-group}
  {A.1}{Proof of Theorem~\ref{thm:group-support}}
\apptocsubsection{app:proof-temporal}
  {A.2}{Proofs of Theorem~\ref{thm:outgoing-temporal-support} and Corollary~\ref{cor:incoming-temporal-support}}
\apptocsubsection{app:proof-common}
  {A.3}{Proof of Lemma~\ref{lem:common-support-alignment}}
\apptocsubsection{app:proof-joint}
  {A.4}{Proof of Theorem~\ref{thm:joint-roles}}
\apptocsubsection{app:common-vs-global}
  {A.5}{Common structural alignment and global permutation invariance}

\vspace{2pt}

\apptocsection{app:extensions}{B}{Extensions of the Temporal Model}
\apptocsubsection{app:higher-lags}
  {B.1}{Higher-order temporal dependencies}
\apptocsubsection{app:instantaneous}
  {B.2}{Instantaneous dependencies}

\vspace{2pt}

\apptocsection{sec:related-work}{C}{Related Work}

\vspace{2pt}

\apptocsection{app:exp}{D}{Additional Experimental Details}
\apptocsubsection{app:synthetic-setup}
  {D.1}{Additional Details for Synthetic Experiments}
\apptocsubsection{app:ami_details}
  {D.2}{Additional Details for AMI Multi-party Meeting Experiment}

\makeatother
\endgroup







\section{Proofs}
\label{app:proofs}

\subsection{Proof of Theorem~\ref{thm:group-support}}
\label{app:proof-group}

\begin{theorem}[Identifiability of grouped observation support]
\label{app-thm:group-support}
For $t\in[T]$, suppose that $\mathcal G$ and
$\widehat{\mathcal G}$ are observationally equivalent and that:
\begin{enumerate}
\renewcommand{\labelenumi}{\roman{enumi}.}
 \item
 The maps $g_t:\mathcal Z_t\to\mathcal X_t$ and
 $\widehat g_t:\widehat{\mathcal Z}_t\to\mathcal X_t$ are $C^1$
 embeddings onto the same observation manifold. Consequently,
 $h_t=\widehat g_t^{-1}\circ g_t$ is a $C^1$ diffeomorphism.

 \item
 There exists $z_t^\star\in\mathcal Z_t$ s.t., for every
 $k\in[K]$,
 $\operatorname{span} \bigl\{(J_{g_t}(z_t^\star))_{r,:}:r\in I_k \bigr\} = \mathbb R^N_{\mathcal S_{t,k}^{\mathbb A}}.$

 \item
 The learned grouped supports are no denser overall than the true grouped supports:$\sum_{k=1}^K \left| \widehat{\mathcal S}_{t,k}^{\mathbb A} \right| \leq \sum_{k=1}^K \left| \mathcal S_{t,k}^{\mathbb A} \right|.$
\end{enumerate}

Let $\widehat z_t^\star=h_t(z_t^\star)$ and
$D_t^\star:=J_{h_t^{-1}}(\widehat z_t^\star)$. Then $(D_t^\star)_{i,v}=0, \quad \forall i\in\mathcal S_{t,k}^{\mathbb A},\;
\forall v\notin\widehat{\mathcal S}_{t,k}^{\mathbb A},\ k\in[K].$
Moreover, every permutation $\sigma_t$ s.t.
$(D_t^\star)_{i,\sigma_t(i)}\neq0 \forall i\in[N]$
aligns the group supports:
\begin{equation*}
\label{app-eq:group-support-identification}
 \widehat{\mathcal S}_{t,k}^{\mathbb A}
 =
 \sigma_t\bigl(\mathcal S_{t,k}^{\mathbb A}\bigr),
 \quad \forall k\in[K].
\end{equation*}
At least one such permutation exists.
\end{theorem}

\begin{proof}
By assumption (i), $h_t:\mathcal Z_t\to\widehat{\mathcal Z}_t$ is a
$C^1$ diffeomorphism. Hence, for every
$\widehat z_t\in\widehat{\mathcal Z}_t$, with
$z_t=h_t^{-1}(\widehat z_t)$,
$\widehat g_t(\widehat z_t) = g_t\left(h_t^{-1}(\widehat z_t)\right).$
Define
$D_t(\widehat z_t):=J_{h_t^{-1}}(\widehat z_t)$.
Applying the chain rule gives:
\begin{equation}
J_{\widehat g_t}(\widehat z_t)
=
J_{g_t}(z_t)D_t(\widehat z_t),
\quad
z_t=h_t^{-1}(\widehat z_t).
\label{eq:app-A-chain-rule}
\end{equation}

At the point in assumption (ii), let
$\widehat z_t^\star=h_t(z_t^\star)$ and
$D_t^\star=D_t(\widehat z_t^\star)$. Since $h_t^{-1}$ is a diffeomorphism, $D_t^\star$ is nonsingular.

We first establish the support separation. Fix an arbitrary group
$k\in[K]$, an index $i\in\mathcal S_{t,k}^{\mathbb A}$, and an index
$v\notin\widehat{\mathcal S}_{t,k}^{\mathbb A}$.
By the definition of the learned group support,
$v\notin\widehat{\mathcal S}_{t,k}^{\mathbb A}$ means that no observation
coordinate indexed by $I_k$ depends on the learned latent coordinate $v$.
Equivalently,
\begin{equation}
\left( J_{\widehat g_t}(\widehat z_t) \right)_{r,v} = 0, \quad \forall r\in I_k, \;
\forall \widehat z_t\in\widehat{\mathcal Z}_t.
\label{eq:app-A-learned-zero}
\end{equation}
In particular, Eq.~\ref{eq:app-A-learned-zero} holds at
$\widehat z_t^\star$.

Since $i\in\mathcal S_{t,k}^{\mathbb A}$, the standard row vector $e_i^\top$ belongs to
$\mathbb R^N_{\mathcal S_{t,k}^{\mathbb A}}$.
Assumption (ii) therefore implies that there exist coefficients
$\{\alpha_r\}_{r\in I_k}$ s.t.
\begin{equation}
e_i^\top
=
\sum_{r\in I_k} \alpha_r \left( J_{g_t}(z_t^\star) \right)_{r,:}.
\label{eq:app-A-span}
\end{equation}

Right-multiplying Eq.~\ref{eq:app-A-span} by $D_t^\star$ yields:
\begin{align}
e_i^\top D_t^\star
&=
\sum_{r\in I_k}
\alpha_r
\left(
J_{g_t}(z_t^\star)
\right)_{r,:}
D_t^\star \notag
\\
&=
\sum_{r\in I_k}
\alpha_r
\left(
J_{\widehat g_t}(\widehat z_t^\star)
\right)_{r,:},
\label{eq:app-A-transform-span}
\end{align}
where the second equality follows from
Eq.~\ref{eq:app-A-chain-rule}.

Taking the $v$-th coordinate of
Eq.~\ref{eq:app-A-transform-span} gives
$(D_t^\star)_{i,v} = \sum_{r\in I_k} \alpha_r \left(J_{\widehat g_t}(\widehat z_t^\star)\right)_{r,v} = 0$.
Since $k$, $i$, and $v$ were arbitrary,
$(D_t^\star)_{i,v}= 0,\quad\forall k\in[K],\;
\forall i\in\mathcal S_{t,k}^{\mathbb A},\;
\forall v\notin\widehat{\mathcal S}_{t,k}^{\mathbb A}.
\label{eq:app-A-separation}$
Equivalently,
\begin{equation}
\operatorname{supp} \left( (D_t^\star)_{i,:}\right)\subseteq
\widehat{\mathcal S}_{t,k}^{\mathbb A},\quad \forall k\in[K],\;
\forall i\in\mathcal S_{t,k}^{\mathbb A}.
\label{eq:app-A-row-support}
\end{equation}

We next show that at least one permutation is supported by the nonzero
pattern of $D_t^\star$. Since $D_t^\star$ is nonsingular,
$\det(D_t^\star)\neq0$. By the Leibniz formula,
\begin{equation}
\det(D_t^\star) = \sum_{\sigma\in S_N}
\operatorname{sgn}(\sigma) \prod_{i=1}^N
(D_t^\star)_{i,\sigma(i)} \neq0,
\label{eq:app-A-leibniz}
\end{equation}
where $S_N$ denotes the set of permutations of $[N]$.
Therefore, at least one term in the sum is nonzero. Hence there exists a
permutation $\sigma_t:[N]\to[N]$ s.t.
\begin{equation}
(D_t^\star)_{i,\sigma_t(i)} \neq0,\quad \forall i\in[N].
\label{eq:app-A-supported-permutation}
\end{equation}

Now fix any permutation $\sigma_t$ satisfying
Eq.~\ref{eq:app-A-supported-permutation}. We show that this permutation aligns every group support. Fix $k\in[K]$ and $i\in\mathcal S_{t,k}^{\mathbb A}$. By
Eq.~\ref{eq:app-A-row-support}, every nonzero entry in row $i$ of
$D_t^\star$ has its column index in
$\widehat{\mathcal S}_{t,k}^{\mathbb A}$. Since $(D_t^\star)_{i,\sigma_t(i)}\neq0$, it follows that
$\sigma_t(i)
\in
\widehat{\mathcal S}_{t,k}^{\mathbb A}.$
Thus,
\begin{equation}
\sigma_t \left( \mathcal S_{t,k}^{\mathbb A} \right) \subseteq
\widehat{\mathcal S}_{t,k}^{\mathbb A}, \quad \forall k\in[K].
\label{eq:app-A-support-inclusion}
\end{equation}

Since $\sigma_t$ is a permutation and hence injective,
\begin{equation}
\left| \mathcal S_{t,k}^{\mathbb A}
\right| = \left| \sigma_t \left( \mathcal S_{t,k}^{\mathbb A} \right) \right| \leq \left| \widehat{\mathcal S}_{t,k}^{\mathbb A} \right|, \quad \forall k\in[K].
\label{eq:app-A-cardinality-lower}
\end{equation}
Summing Eq.~\ref{eq:app-A-cardinality-lower} over $k$ gives:
\begin{equation}
\sum_{k=1}^K \left| \mathcal S_{t,k}^{\mathbb A} \right|
\leq
\sum_{k=1}^K \left| \widehat{\mathcal S}_{t,k}^{\mathbb A} \right|.
\label{eq:app-A-total-lower}
\end{equation}
Therefore, by assumption (iii),
\begin{equation}
\sum_{k=1}^K \left| \widehat{\mathcal S}_{t,k}^{\mathbb A} \right| = \sum_{k=1}^K \left|
\mathcal S_{t,k}^{\mathbb A} \right|.
\label{eq:app-A-total-equality}
\end{equation}

For each $k\in[K]$, define $ \Delta_k := \left| \widehat{\mathcal S}_{t,k}^{\mathbb A} \right| - \left| \mathcal S_{t,k}^{\mathbb A} \right|. $

By Eq.~\ref{eq:app-A-cardinality-lower},
$\Delta_k\geq0$ for every $k$, while
Eq.~\ref{eq:app-A-total-equality} implies
$\sum_{k=1}^K\Delta_k=0$. Hence $\Delta_k=0$ for every $k\in[K]$, i.e.,
\begin{equation}
\left|
\widehat{\mathcal S}_{t,k}^{\mathbb A}
\right|
=
\left|
\mathcal S_{t,k}^{\mathbb A}
\right|,
\quad
\forall k\in[K].
\label{eq:app-A-cardinality-equality}
\end{equation}

Combining Eq.~\ref{eq:app-A-support-inclusion} with
Eq.~\ref{eq:app-A-cardinality-equality} yields:
$$
\widehat{\mathcal S}_{t,k}^{\mathbb A}
=
\sigma_t
\left(
\mathcal S_{t,k}^{\mathbb A}
\right),
\quad
\forall k\in[K].
$$

The permutation $\sigma_t$ was chosen once from the single nonsingular matrix $D_t^\star$, and the argument above holds simultaneously for every $k\in[K]$. Therefore, the same permutation aligns all group-specific observation supports. Since the argument applies to any permutation satisfying Eq.~\ref{eq:app-A-supported-permutation}, every such permutation has the claimed property.

\end{proof}

\subsection{Proofs of Theorem~\ref{thm:outgoing-temporal-support} and Corollary~\ref{cor:incoming-temporal-support}}
\label{app:proof-temporal}

\begin{theorem}[Identifiability of outgoing temporal support]
\label{thm:app-outgoing-temporal-support}
For $t\in\{1,\ldots,T-1\}$, suppose that the true model $\mathcal G$ and learned model $\widehat{\mathcal G}$ are
observationally equivalent and that:
\begin{enumerate}
\renewcommand{\labelenumi}{\roman{enumi}.}
 \item
 The transition densities $p(z_{t+1}\mid z_t), p(\widehat z_{t+1}\mid\widehat z_t)$ are positive and $C^2$, and the observation embeddings induce $C^1$ diffeomorphisms $h_t:\mathcal Z_t\to\widehat{\mathcal Z}_t$ and $h_{t+1}:\mathcal Z_{t+1}\to\widehat{\mathcal Z}_{t+1}$.

 \item
 The true and learned delayed graphs are faithful to the cross-Hessian. For every $i,j\in[N]$
 $$
 (G_{t+1}^d)_{j,i}=1
 \iff
 \exists (z_{t+1},z_t)\in \mathcal Z_{t+1}\times\mathcal Z_t,
 \;\; \text{s.t.}\;
 [M_t(z_{t+1},z_t)]_{j,i}\neq0,
 $$
 with the analogous equivalence holding for $\widehat G_{t+1}^d$ and $\widehat M_t$.

 \item
 For $ \forall \;z_t^\star\in\mathcal Z_t$, $\operatorname{span}
 \bigl\{ M_t(z_{t+1},z_t^\star): z_{t+1}\in\mathcal Z_{t+1} \bigr\} = \mathbb R^{N\times N}_{G_{t+1}^d}.$

 \item
 The learned outgoing support satisfies
 $\lvert\widehat{\mathcal S}_t^{\mathbb C}\rvert
 \leq\lvert\mathcal S_t^{\mathbb C}\rvert$.
\end{enumerate}
Then, for $ \forall \;\widehat z_t\in\widehat{\mathcal Z}_t$,
$ \frac{\partial Z_{t,i}}{\partial\widehat Z_{t,v}}=0 \quad \text{for } i\in\mathcal S_t^{\mathbb C},\ v\notin\widehat{\mathcal S}_t^{\mathbb C}.$ For any $\widehat z_t^0$, let
$D_t^0:=J_{h_t^{-1}}(\widehat z_t^0)$. At least one permutation $\sigma_t$ exists s.t. $(D_t^0)_{i,\sigma_t(i)}\neq0$ for $ \forall \;i\in[N]$ satisfies $\widehat{\mathcal S}_t^{\mathbb C} = \sigma_t\bigl(\mathcal S_t^{\mathbb C}\bigr)$.
\end{theorem}

\begin{proof}
For $\widehat z_t\in\widehat{\mathcal Z}_t$ and
$\widehat z_{t+1}\in\widehat{\mathcal Z}_{t+1}$, let $z_t=h_t^{-1}(\widehat z_t), \quad z_{t+1}=h_{t+1}^{-1}(\widehat z_{t+1}),$
and define:
$$
A_t(\widehat z_{t+1}):=J_{h_{t+1}^{-1}}(\widehat z_{t+1}), \quad D_t(\widehat z_t) := J_{h_t^{-1}}(\widehat z_t).
$$
By the change-of-variables formula for conditional densities,
\begin{equation}
\widehat p(\widehat z_{t+1}\mid\widehat z_t) = \frac{\widehat p(\widehat z_{t+1},\widehat z_t)}{\widehat p(\widehat z_t)}
=
p(z_{t+1}\mid z_t)
\left|
\det A_t(\widehat z_{t+1})
\right|.
\label{eq:app-outgoing-change-var}
\end{equation}
Let $q_{t+1}:=\log p$ and $\widehat q_{t+1}:=\log\widehat p$. Taking logarithms in Eq.~\ref{eq:app-outgoing-change-var} gives
\begin{equation}
\widehat q_{t+1}(\widehat z_{t+1}\mid\widehat z_t) = q_{t+1}(z_{t+1}\mid z_t) + \log \left| \det A_t(\widehat z_{t+1}) \right|.
\label{eq:app-outgoing-log-change-var}
\end{equation}
The second term in Eq.~\ref{eq:app-outgoing-log-change-var} depends only on $\widehat z_{t+1}$ and therefore vanishes after taking a derivative with respect to $\widehat z_t$.

We now derive the transformation rule for the cross-Hessian. For
$r,v\in[N]$, the chain rule gives:
\begin{align*}
[\widehat M_t(\widehat z_{t+1},\widehat z_t)]_{r,v}
&=
\frac{\partial^2
\widehat q_{t+1}(\widehat z_{t+1}\mid\widehat z_t)}
{\partial\widehat z_{t+1,r}\partial\widehat z_{t,v}} \\
&=
\sum_{j=1}^N
\sum_{i=1}^N
\frac{\partial z_{t+1,j}}
{\partial\widehat z_{t+1,r}}
\frac{\partial^2
q_{t+1}(z_{t+1}\mid z_t)}
{\partial z_{t+1,j}\partial z_{t,i}}
\frac{\partial z_{t,i}}
{\partial\widehat z_{t,v}}.
\end{align*}
Hence,
\begin{equation}
\widehat M_t(\widehat z_{t+1},\widehat z_t) = A_t(\widehat z_{t+1})^{\top} M_t(z_{t+1},z_t) D_t(\widehat z_t),
\label{eq:app-outgoing-cross-transform}
\end{equation}
where both
$A_t(\widehat z_{t+1})$ and $D_t(\widehat z_t)$ are nonsingular by assumption~(i).

For any $i\in\mathcal S_t^{\mathbb C}$ and an arbitrary $z_t^\star\in\mathcal Z_t$, since $i$ is outgoing-active, there exists $j\in[N]$ s.t. $(G_{t+1}^d)_{j,i}=1$. Therefore, the coordinate matrix $ E_{ji}:=e_je_i^\top $ belongs to $\mathbb R_{G_{t+1}^d}^{N\times N}$. By assumption~(iii), there exist $L\in\mathbb N$, points $\{z_{t+1}^{(\ell)}\}_{\ell=1}^L\subseteq\mathcal Z_{t+1}$, and coefficients
$\{\alpha_\ell\}_{\ell=1}^L$ such that:
\begin{equation}
E_{ji}
=
\sum_{\ell=1}^L
\alpha_\ell
M_t
\left(
z_{t+1}^{(\ell)},z_t^\star
\right).
\label{eq:app-outgoing-span}
\end{equation}
Let $\widehat z_t^\star=h_t(z_t^\star)$ and $D_t^\star:=D_t(\widehat z_t^\star)$. For every $\ell\in[L]$, define:
$$
\widehat z_{t+1}^{(\ell)} := h_{t+1}
\left( z_{t+1}^{(\ell)} \right), \quad
A_\ell := A_t \left( \widehat z_{t+1}^{(\ell)} \right),
$$
$$ 
M_\ell := M_t \left( z_{t+1}^{(\ell)},z_t^\star \right), \quad \widehat M_\ell := \widehat M_t
\left( \widehat z_{t+1}^{(\ell)},\widehat z_t^\star \right).
$$
Eq.~\ref{eq:app-outgoing-cross-transform} gives $ \widehat M_\ell = A_\ell^\top M_\ell D_t^\star. $ Since $A_\ell$ is nonsingular,
\begin{equation}
M_\ell D_t^\star = A_\ell^{-\top}
\widehat M_\ell.
\label{eq:app-outgoing-left-invert}
\end{equation}
Right-multiplying Eq.~\ref{eq:app-outgoing-span} by $D_t^\star$ and applying Eq.~\ref{eq:app-outgoing-left-invert} yields:
\begin{equation}
E_{ji}D_t^\star = \sum_{\ell=1}^L
\alpha_\ell A_\ell^{-\top} \widehat M_\ell. 
\label{eq:app-outgoing-propagate}
\end{equation}
Now take any
$v\notin\widehat{\mathcal S}_t^{\mathbb C}$. We have $ (\widehat G_{t+1}^d)_{r,v}=0, \; \forall r\in[N]. $
Assumption~(ii) therefore implies that the $v$-th column of the learned cross-Hessian vanishes everywhere:
\begin{equation}
[\widehat M_t(\widehat z_{t+1},\widehat z_t)]_{:,v}
= 0, \quad \forall (\widehat z_{t+1},\widehat z_t) \in \widehat{\mathcal Z}_{t+1} \times \widehat{\mathcal Z}_t.
\label{eq:app-outgoing-learned-zero}
\end{equation}
In particular, $(\widehat M_\ell)_{:,v}=0$ for every $\ell\in[L]$. Therefore,
$$
(A_\ell^{-\top}\widehat M_\ell)_{:,v} = 0, \quad \forall \ell\in[L].
$$
Taking the $v$-th column of Eq.~\ref{eq:app-outgoing-propagate} gives:
\begin{equation}
(E_{ji}D_t^\star)_{:,v} = 0.
\label{eq:app-outgoing-column-zero}
\end{equation}
Since $E_{ji}D_t^\star = e_j(e_i^\top D_t^\star),$ its $v$-th column is $ (E_{ji}D_t^\star)_{:,v} = e_j(D_t^\star)_{i,v}.$
Combining this identity with Eq.~\ref{eq:app-outgoing-column-zero} yields $(D_t^\star)_{i,v} = 0$.

The point $z_t^\star$ was arbitrary in $\mathcal Z_t$. Since $h_t$ is a bijection between $\mathcal Z_t$ and $\widehat{\mathcal Z}_t$, we obtain:
\begin{equation} [J_{h_t^{-1}}(\widehat z_t)]_{i,v} = 0, \quad \forall \widehat z_t\in\widehat{\mathcal Z}_t, \; \forall i\in\mathcal S_t^{\mathbb C}, \; \forall v\notin\widehat{\mathcal S}_t^{\mathbb C}. \label{eq:app-outgoing-separation} \end{equation}
Equivalently,
\begin{equation}
\operatorname{supp} \left( J_{h_t^{-1}}(\widehat z_t)_{i,:}
\right) \subseteq \widehat{\mathcal S}_t^{\mathbb C}, \quad \forall \widehat z_t\in\widehat{\mathcal Z}_t, \; \forall i\in\mathcal S_t^{\mathbb C}.
\label{eq:app-outgoing-row-support}
\end{equation}

We next identify the outgoing support up to permutation. For $\widehat z_t^0\in\widehat{\mathcal Z}_t$ and let $D_t^0:=J_{h_t^{-1}}(\widehat z_t^0)$. Since $h_t^{-1}$ is a
diffeomorphism, $D_t^0$ is nonsingular and $\det(D_t^0)\neq0$. By the Leibniz formula,
\begin{equation}
\det(D_t^0) = \sum_{\sigma\in S_N}
\operatorname{sgn}(\sigma) \prod_{i=1}^N (D_t^0)_{i,\sigma(i)}
\neq0.
 \label{eq:app-outgoing-leibniz}
\end{equation}
Therefore, at least one permutation $\sigma_t\in S_N$ satisfies:
\begin{equation}
(D_t^0)_{i,\sigma_t(i)} \neq0, \quad \forall i\in[N].
\label{eq:app-outgoing-supported-permutation}
\end{equation}

Now fix any permutation $\sigma_t$ satisfying Eq.~\ref{eq:app-outgoing-supported-permutation}. For every $i\in\mathcal S_t^{\mathbb C}$, Eq.~\ref{eq:app-outgoing-row-support} evaluated at $\widehat z_t^0$ implies that every nonzero entry in row $i$ of $D_t^0$ has its column index in $\widehat{\mathcal S}_t^{\mathbb C}$. Since $(D_t^0)_{i,\sigma_t(i)}\neq0$, we have $\sigma_t(i) \in \widehat{\mathcal S}_t^{\mathbb C}.$
Thus,
\begin{equation}
\sigma_t \left(\mathcal S_t^{\mathbb C} \right) \subseteq \widehat{\mathcal S}_t^{\mathbb C}.
\label{eq:app-outgoing-inclusion}
\end{equation}

Since $\sigma_t$ is a permutation, $\left| \sigma_t \left( \mathcal S_t^{\mathbb C} \right) \right| = \left| \mathcal S_t^{\mathbb C} \right|$. Eq.~\ref{eq:app-outgoing-inclusion} therefore gives: 
$$ |\mathcal S_t^{\mathbb C}| \leq
|\widehat{\mathcal S}_t^{\mathbb C}|. 
$$ 
Combining this inequality with assumption~(iv) yields: 
$$
\widehat{\mathcal S}_t^{\mathbb C} = \sigma_t \left( \mathcal S_t^{\mathbb C} \right).
$$
The argument applies to every permutation supported by the nonzero pattern of $D_t^0$, and Eq.~\ref{eq:app-outgoing-leibniz} guarantees that at least one such permutation exists.
\end{proof}

\begin{corollary}[Identifiability of incoming temporal support]
\label{cor:app-incoming-temporal-support}
For $t\in\{2,\ldots,T\}$, suppose that the true model $\mathcal G$ and learned model $\widehat{\mathcal G}$ are
observationally equivalent and that:
\begin{enumerate}
\renewcommand{\labelenumi}{\roman{enumi}.}
 \item
 The smoothness and faithfulness conditions in
 Theorem~\ref{thm:outgoing-temporal-support} hold for $t-1 \to t$.
 
 \item
 For $ \forall \;z_t^\star\in\mathcal Z_t$,$ \operatorname{span} \bigl\{M_{t-1}(z_t^\star,z_{t-1}): z_{t-1}\in\mathcal Z_{t-1} \bigr\} = \mathbb R^{N\times N}_{G_t^d}$.

 \item
 The learned incoming support satisfies $\lvert\widehat{\mathcal S}_t^{\mathbb B}\rvert \leq\lvert\mathcal S_t^{\mathbb B}\rvert$.
\end{enumerate}
Then, for $ \forall \;\widehat z_t$, $ \frac{\partial Z_{t,i}}{\partial\widehat Z_{t,v}}=0 \quad \text{for } i\in\mathcal S_t^{\mathbb B},\ v\notin\widehat{\mathcal S}_t^{\mathbb B}$.
For any $\widehat z_t^0$, At least one permutation $\sigma_t$ exists, supported by the nonzero pattern of $D_t^0=J_{h_t^{-1}}(\widehat z_t^0)$ satisfies $\widehat{\mathcal S}_t^{\mathbb B} = \sigma_t\bigl(\mathcal S_t^{\mathbb B}\bigr).$
\end{corollary}

\begin{proof}
The result follows from the row-side counterpart of the argument for Theorem~\ref{thm:outgoing-temporal-support}. For the transition from $t-1$ to $t$, the cross-Hessian transforms as:
$$
\widehat M_{t-1}(\widehat z_t,\widehat z_{t-1}) = J_{h_t^{-1}}(\widehat z_t)^\top M_{t-1}(z_t,z_{t-1}) J_{h_{t-1}^{-1}}(\widehat z_{t-1}).
$$
Theorem~\ref{thm:outgoing-temporal-support} uses the column support of the cross-Hessian and right multiplication by $J_{h_t^{-1}}$ to identify outgoing-active coordinates. Here, we exchange the row and column roles.

For $z_t^\star$ and vary $z_{t-1}$, Assumption~(ii) allows every coordinate matrix $E_{ij}=e_i e_j^\top$ associated with an edge
$(G_t^d)_{i,j}=1$ to be expressed as a linear combination of $M_{t-1}(z_t^\star,z_{t-1})$. Since $J_{h_t^{-1}}(h_t(z_t^\star))$ is fixed while $z_{t-1}$ varies, multiplying
$\widehat M_{t-1}(\widehat z_t,\widehat z_{t-1})$ on the right by the inverse of $J_{h_{t-1}^{-1}}(\widehat z_{t-1})$ shows that
$$
[J_{h_t^{-1}}(\widehat z_t)]_{i,v} = 0,
\quad \forall i\in\mathcal S_t^{\mathbb B}, \; \forall v\notin\widehat{\mathcal S}_t^{\mathbb B}.
$$
The same supported-permutation argument as in Theorem~\ref{thm:outgoing-temporal-support}, followed by the cardinality condition in assumption~(iii), gives $
\widehat{\mathcal S}_t^{\mathbb B} = \sigma_t \left( \mathcal S_t^{\mathbb B} \right).$
\end{proof}

\subsection{Proof of Lemma~\ref{lem:common-support-alignment}}
\label{app:proof-common}

\begin{lemma}[Common support alignment]
\label{lem:app-common-support-alignment}
For $t\in\{2,\ldots,T-1\}$, suppose that the assumptions of
Theorem~\ref{thm:group-support},
Theorem~\ref{thm:outgoing-temporal-support}, and
Corollary~\ref{cor:incoming-temporal-support} hold. Let $z_t^\star\in\mathcal Z_t$ be the point in assumption~(ii) of Theorem~\ref{thm:group-support}, and define $\widehat z_t^\star:=h_t(z_t^\star)$ and $D_t^\star:=J_{h_t^{-1}}(\widehat z_t^\star)$. There exists at least one permutation $\sigma_t:[N]\to[N]$ s.t.
$(D_t^\star)_{i,\sigma_t(i)}\neq 0 \;$ for $\; \forall \;i\in[N], \; \forall \; k\in[K]$ simultaneously satisfies:
\begin{align}
 \widehat{\mathcal S}_{t,k}^{\mathbb A} = \sigma_t\bigl(\mathcal S_{t,k}^{\mathbb A}\bigr),\;\;
 \widehat{\mathcal S}_t^{\mathbb B} = \sigma_t\bigl(\mathcal S_t^{\mathbb B}\bigr), \;\;
 \widehat{\mathcal S}_t^{\mathbb C} = \sigma_t\bigl(\mathcal S_t^{\mathbb C}\bigr).
\end{align}
\end{lemma}

\begin{proof}
Since $h_t$ is a diffeomorphism, $D_t^\star=J_{h_t^{-1}}(\widehat z_t^\star)$ is nonsingular. Hence $\det(D_t^\star)\neq0$. By the Leibniz formula,
\begin{equation*}
\det(D_t^\star) = \sum_{\sigma\in S_N} \operatorname{sgn}(\sigma) \prod_{i=1}^N (D_t^\star)_{i,\sigma(i)} \neq0,
\label{eq:app-common-leibniz}
\end{equation*}
where $S_N$ denotes the set of permutations of $[N]$. Therefore, at least one term in the sum is nonzero, and there exists a permutation $\sigma_t:[N]\to[N]$ s.t.
\begin{equation}
(D_t^\star)_{i,\sigma_t(i)} \neq0, \quad
\forall i\in[N].
\label{eq:app-common-supported-permutation}
\end{equation}
We show that every permutation satisfying Eq.~\ref{eq:app-common-supported-permutation} simultaneously aligns the
three support structures. First, consider the grouped observation supports. By Theorem~\ref{thm:group-support},
\begin{equation}
\operatorname{supp} \left( (D_t^\star)_{i,:} \right) \subseteq \widehat{\mathcal S}_{t,k}^{\mathbb A}, \quad \forall k\in[K],\; \forall i\in\mathcal S_{t,k}^{\mathbb A}.
\label{eq:app-common-A-separation}
\end{equation}
Fix $k\in[K]$ and $i\in\mathcal S_{t,k}^{\mathbb A}$. Eqs.~\ref{eq:app-common-supported-permutation} and
\ref{eq:app-common-A-separation} imply $ \sigma_t(i) \in \widehat{\mathcal S}_{t,k}^{\mathbb A}.$ Hence,
\begin{equation}
\sigma_t \left( \mathcal S_{t,k}^{\mathbb A} \right) \subseteq \widehat{\mathcal S}_{t,k}^{\mathbb A}, \quad \forall k\in[K].
\label{eq:app-common-A-inclusion}
\end{equation}
Since $\sigma_t$ is injective,
\begin{equation}
\left| \mathcal S_{t,k}^{\mathbb A} \right| = \left| \sigma_t \left( \mathcal S_{t,k}^{\mathbb A} \right) \right| \leq \left| \widehat{\mathcal S}_{t,k}^{\mathbb A} \right|, \quad \forall k\in[K].
\label{eq:app-common-A-cardinality}
\end{equation}
\begin{equation}
\sum_{k=1}^K \left| \mathcal S_{t,k}^{\mathbb A} \right| \leq \sum_{k=1}^K \left| \widehat{\mathcal S}_{t,k}^{\mathbb A} \right|.
\label{eq:app-common-A-total-lower}
\end{equation}
The total group-incidence sparsity assumption in
Theorem~\ref{thm:group-support} gives the reverse inequality, so:
\begin{equation}
\sum_{k=1}^K \left| \widehat{\mathcal S}_{t,k}^{\mathbb A} \right| = \sum_{k=1}^K \left| \mathcal S_{t,k}^{\mathbb A} \right|.
\label{eq:app-common-A-total-equality}
\end{equation}
Every difference $\left| \widehat{\mathcal S}_{t,k}^{\mathbb A} \right| - \left| \mathcal S_{t,k}^{\mathbb A} \right|$
is nonnegative by Eq.~\ref{eq:app-common-A-cardinality}, while their sum is zero by Eq.~\ref{eq:app-common-A-total-equality}. Therefore, $\left| \widehat{\mathcal S}_{t,k}^{\mathbb A} \right| = \left| \mathcal S_{t,k}^{\mathbb A} \right|, \; \forall k\in[K].$ Combining this equality with Eq.~\ref{eq:app-common-A-inclusion} yields:
\begin{equation}
\widehat{\mathcal S}_{t,k}^{\mathbb A} = \sigma_t \left( \mathcal S_{t,k}^{\mathbb A} \right), \quad \forall k\in[K].
\label{eq:app-common-A-equality}
\end{equation}
Next, we consider the incoming temporal support. By Corollary~\ref{cor:incoming-temporal-support}, for every $\widehat z_t\in\widehat{\mathcal Z}_t$, $\operatorname{supp} \left( J_{h_t^{-1}}(\widehat z_t)_{i,:} \right) \subseteq \widehat{\mathcal S}_t^{\mathbb B}, \; \forall i\in\mathcal S_t^{\mathbb B}$. Evaluating this relation at $\widehat z_t^\star$ gives
\begin{equation}
\operatorname{supp} \left( (D_t^\star)_{i,:} \right) \subseteq \widehat{\mathcal S}_t^{\mathbb B}, \quad \forall i\in\mathcal S_t^{\mathbb B}.
\label{eq:app-common-B-separation}
\end{equation}
For every $i\in\mathcal S_t^{\mathbb B}$, Eqs.~\ref{eq:app-common-supported-permutation} and \ref{eq:app-common-B-separation} imply
$\sigma_t(i)\in\widehat{\mathcal S}_t^{\mathbb B}$. Thus,
\begin{equation}
\sigma_t \left( \mathcal S_t^{\mathbb B} \right) \subseteq \widehat{\mathcal S}_t^{\mathbb B}.
\label{eq:app-common-B-inclusion}
\end{equation}
Since $\sigma_t$ is injective, $\left| \mathcal S_t^{\mathbb B} \right| = \left| \sigma_t \left( \mathcal S_t^{\mathbb B} \right) \right| \leq \left| \widehat{\mathcal S}_t^{\mathbb B} \right|.$
The sparsity condition in Corollary~\ref{cor:incoming-temporal-support} gives $\left|
\widehat{\mathcal S}_t^{\mathbb B}\right|\leq\left|\mathcal S_t^{\mathbb B}\right|$. Therefore the two cardinalities are equal, and Eq.~\ref{eq:app-common-B-inclusion} gives:
\begin{equation}
\widehat{\mathcal S}_t^{\mathbb B} = \sigma_t \left( \mathcal S_t^{\mathbb B} \right).
\label{eq:app-common-B-equality}
\end{equation}
Finally, Theorem~\ref{thm:outgoing-temporal-support} establishes that, for every $\widehat z_t\in\widehat{\mathcal Z}_t$, $\operatorname{supp} \left( J_{h_t^{-1}}(\widehat z_t)_{i,:} \right) \subseteq \widehat{\mathcal S}_t^{\mathbb C}, \; \forall i\in\mathcal S_t^{\mathbb C}.$ Evaluating at $\widehat z_t^\star$ gives
\begin{equation}
\operatorname{supp} \left( (D_t^\star)_{i,:} \right) \subseteq \widehat{\mathcal S}_t^{\mathbb C}, \quad \forall i\in\mathcal S_t^{\mathbb C}.
\label{eq:app-common-C-separation}
\end{equation}
Hence, for every $i\in\mathcal S_t^{\mathbb C}$,
Eq.~\ref{eq:app-common-supported-permutation} implies
$\sigma_t(i)\in\widehat{\mathcal S}_t^{\mathbb C}$, and therefore:
\begin{equation}
\sigma_t \left( \mathcal S_t^{\mathbb C} \right) \subseteq \widehat{\mathcal S}_t^{\mathbb C}.
\label{eq:app-common-C-inclusion}
\end{equation}
Injectivity of $\sigma_t$ and the sparsity condition in
Theorem~\ref{thm:outgoing-temporal-support} give $\left| \mathcal S_t^{\mathbb C} \right| = \left| \sigma_t \left( \mathcal S_t^{\mathbb C} \right) \right| \leq \left| \widehat{\mathcal S}_t^{\mathbb C} \right| \leq \left| \mathcal S_t^{\mathbb C} \right|$. All inequalities are therefore equalities, and
Eq.~\ref{eq:app-common-C-inclusion} yields:
\begin{equation}
\widehat{\mathcal S}_t^{\mathbb C} = \sigma_t \left( \mathcal S_t^{\mathbb C} \right).
\label{eq:app-common-C-equality}
\end{equation}
Eqs.~\ref{eq:app-common-A-equality}, \ref{eq:app-common-B-equality}, and \ref{eq:app-common-C-equality} are obtained from the same permutation $\sigma_t$ selected from the nonzero pattern of the single matrix $D_t^\star$. Hence $\sigma_t$ simultaneously aligns the grouped observation, incoming, and outgoing supports. The argument applies to every permutation satisfying Eq.~\ref{eq:app-common-supported-permutation}.
\end{proof}

\subsection{Proof of Theorem~\ref{thm:joint-roles}}
\label{app:proof-joint}
\begin{theorem}[Joint identifiability of temporal latent roles]
\label{thm:app-joint-roles}
Under the conditions of Lemma~\ref{lem:common-support-alignment}, there exists a permutation $\sigma_t:[N]\to[N]$ s.t., for $ \forall \;i\in[N],\; \forall \;k\in[K]$,
\begin{equation}
\label{eq:app-joint-indicators}
 \left( \widehat{\mathbb A}_{t,\sigma_t(i)}^{(k)}, \widehat{\mathbb B}_{t,\sigma_t(i)}, \widehat{\mathbb C}_{t,\sigma_t(i)} \right)
 =
 \left(\mathbb A_{t,i}^{(k)}, \mathbb B_{t,i}, \mathbb C_{t,i} \right).
\end{equation}
Consequently, for every
$\boldsymbol a\in\{0,1\}^K$ and $b,c\in\{0,1\}$,
\begin{equation}
\label{eq:app-joint-role-class-identification}
 \widehat{\mathcal R}_t^{\boldsymbol a,b,c}
 =
 \sigma_t\bigl(\mathcal R_t^{\boldsymbol a,b,c}\bigr).
\end{equation}
\end{theorem}

\begin{proof}
By Lemma~\ref{lem:common-support-alignment}, there exists a single
permutation $\sigma_t:[N]\to[N]$ s.t. for $\forall k \in [K]$:
\begin{align*}
 \widehat{\mathcal S}_{t,k}^{\mathbb A} = \sigma_t\bigl(\mathcal S_{t,k}^{\mathbb A}\bigr),\;\;
 \widehat{\mathcal S}_t^{\mathbb B} = \sigma_t\bigl(\mathcal S_t^{\mathbb B}\bigr), \;\;
 \widehat{\mathcal S}_t^{\mathbb C} = \sigma_t\bigl(\mathcal S_t^{\mathbb C}\bigr).
\end{align*}

For every $i\in[N]$ and $k\in[K]$,
\begin{align*}
\widehat{\mathbb A}_{t,\sigma_t(i)}^{(k)}=1
&\iff
\sigma_t(i)\in\widehat{\mathcal S}_{t,k}^{\mathbb A}
\\
&\iff
\sigma_t(i)\in
\sigma_t
\left(
\mathcal S_{t,k}^{\mathbb A}
\right)
\\
&\iff
i\in\mathcal S_{t,k}^{\mathbb A}
\\
&\iff
\mathbb A_{t,i}^{(k)}=1.
\end{align*}
Since both quantities are binary,
\begin{equation}
\widehat{\mathbb A}_{t,\sigma_t(i)}^{(k)}
=
\mathbb A_{t,i}^{(k)},
\quad
\forall i\in[N],
\;
\forall k\in[K].
\label{eq:app-joint-A}
\end{equation}

For the incoming indicator, for every $i\in[N]$,
\begin{align*}
\widehat{\mathbb B}_{t,\sigma_t(i)}=1
&\iff
\sigma_t(i)\in\widehat{\mathcal S}_t^{\mathbb B}
\\
&\iff
\sigma_t(i)\in
\sigma_t
\left(
\mathcal S_t^{\mathbb B}
\right)
\\
&\iff
i\in\mathcal S_t^{\mathbb B}
\\
&\iff
\mathbb B_{t,i}=1.
\end{align*}
Hence,
\begin{equation}
\widehat{\mathbb B}_{t,\sigma_t(i)}
=
\mathbb B_{t,i},
\quad
\forall i\in[N].
\label{eq:app-joint-B}
\end{equation}

Similarly, for the outgoing indicator,
\begin{align*}
\widehat{\mathbb C}_{t,\sigma_t(i)}=1
&\iff
\sigma_t(i)\in\widehat{\mathcal S}_t^{\mathbb C}
\\
&\iff
\sigma_t(i)\in
\sigma_t
\left(
\mathcal S_t^{\mathbb C}
\right)
\\
&\iff
i\in\mathcal S_t^{\mathbb C}
\\
&\iff
\mathbb C_{t,i}=1.
\end{align*}
Therefore,
\begin{equation}
\widehat{\mathbb C}_{t,\sigma_t(i)}
=
\mathbb C_{t,i},
\quad
\forall i\in[N].
\label{eq:app-joint-C}
\end{equation}

Eqs.~\ref{eq:app-joint-A}, \ref{eq:app-joint-B}, and \ref{eq:app-joint-C} establish the joint signature equality. For $\mathbf a=(a_1,\ldots,a_K)\in\{0,1\}^K$ and $b,c\in\{0,1\}$. For every
$i\in[N]$,
\begin{align*}
i\in\mathcal R_t^{\mathbf a,b,c} &\iff \mathbb A_{t,i}^{(k)}=a_k \;\text{for all}\; k\in[K], \quad \mathbb B_{t,i}=b, \quad \mathbb C_{t,i}=c
\\
&\iff \widehat{\mathbb A}_{t,\sigma_t(i)}^{(k)}=a_k \;\text{for all}\; k\in[K], \quad \widehat{\mathbb B}_{t,\sigma_t(i)}=b, \quad \widehat{\mathbb C}_{t,\sigma_t(i)}=c
\\
&\iff \sigma_t(i) \in \widehat{\mathcal R}_t^{\mathbf a,b,c}.
\end{align*}
Thus,
$$
i\in\mathcal R_t^{\mathbf a,b,c}
\iff
\sigma_t(i)\in\widehat{\mathcal R}_t^{\mathbf a,b,c}.
$$
Since $\sigma_t$ is bijective, the pointwise membership equivalence implies:
$$
\widehat{\mathcal R}_t^{\mathbf a,b,c}
=
\sigma_t
\left(
\mathcal R_t^{\mathbf a,b,c}
\right).
$$
\end{proof}

\subsection{Common structural alignment and global permutation invariance}
\label{app:common-vs-global}

\begin{figure}[t]
 \centering
 \includegraphics[width=\textwidth]{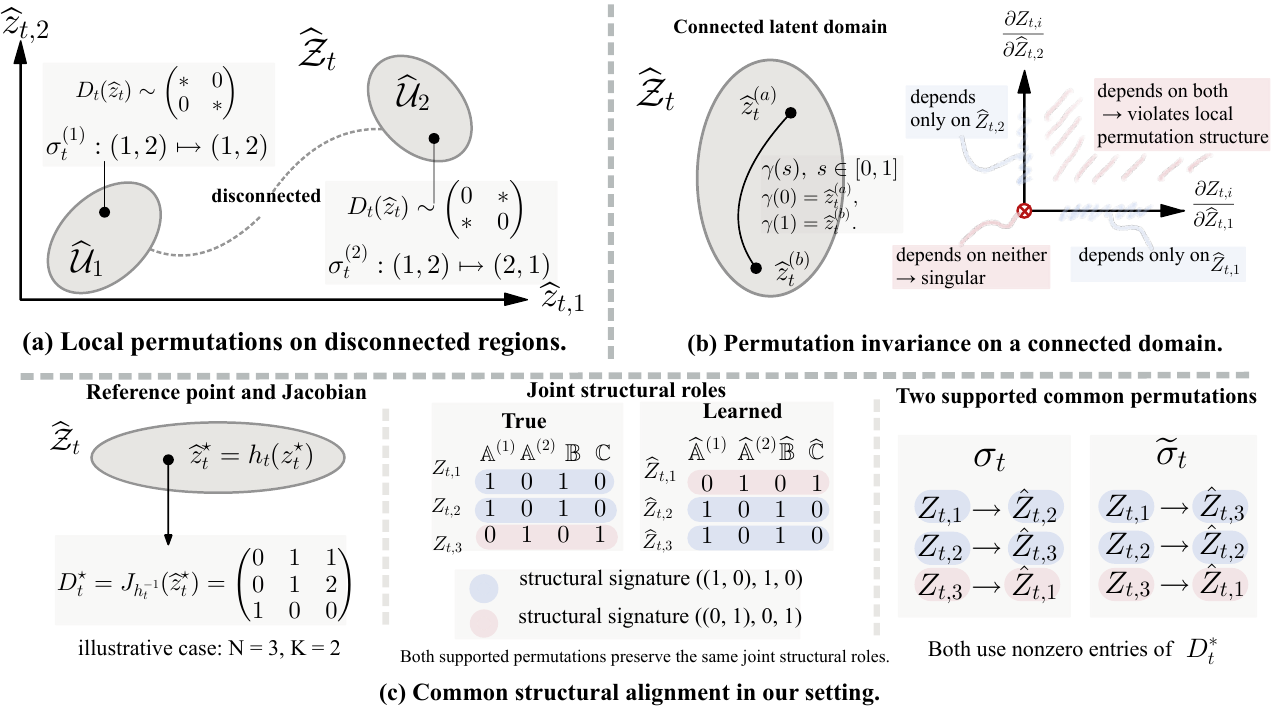}
 \caption{Common structural alignment and global permutation invariance. (a) Disconnected regions may admit different local permutations; (b) continuity on a connected domain prevents permutation switching under the local permutation constraint; (c) in our setting, a supported permutation of $D_t^\star$ aligns the group, incoming, and outgoing structural roles.}
 \label{fig:common-vs-global}
\end{figure}

Lemma~\ref{lem:common-support-alignment} concerns the correspondence among the three structural roles at the same time point. Recall that $D_t^\star=J_{h_t^{-1}}(\widehat z_t^\star)$ is nonsingular. The preceding results show that the group, incoming, and outgoing supports all restrict the rows of this same matrix. Hence, a permutation supported by the nonzero pattern of $D_t^\star$ can align $\mathbb A$, $\mathbb B$, and $\mathbb C$ simultaneously. Figure~\ref{fig:common-vs-global} illustrates how this common alignment differs from the global permutation invariance considered in related identifiability results.

\paragraph{Relation to global permutation invariance.}
Figure~\ref{fig:common-vs-global} (a) illustrates why a locally valid permutation need not remain the same over the whole latent domain. The two disconnected regions $\widehat{\mathcal U}_1$ and $\widehat{\mathcal U}_2$ admit different local Jacobian support patterns: one corresponds to the identity assignment and the other to a coordinate swap. Since there is no continuous path connecting the two regions, these local assignments can differ without contradiction. This type of example appears in the discussion of local and global disentanglement in \citep{lachapelle2023additive}.

Figure~\ref{fig:common-vs-global}(b) gives the geometric intuition for how path connectedness and continuity can prevent such a switch. Consider one row of $J_{h_t^{-1}}(\widehat z_t)$ in two dimensions. Its two entries are:
$$
\left(
\frac{\partial Z_{t,i}}{\partial \widehat Z_{t,1}},
\frac{\partial Z_{t,i}}{\partial \widehat Z_{t,2}}
\right).
$$
If this row has only one nonzero entry, it lies on one of the two coordinate axes. Along the continuous path $\gamma(s)$ from $\widehat z_t^{(a)}$ to $\widehat z_t^{(b)}$, switching from one active coordinate to the other would require the row to pass through either the
origin or a point with two nonzero entries. The origin gives a zero Jacobian row and violates nonsingularity, while the second case violates the assumed local permutation structure. The active coordinate therefore stays fixed along the path. This is the continuity argument used in~\citep{3692070.3692351} to establish permutation invariance, with a related block-level argument developed by~\citep{lachapelle2023additive}.

Our common alignment result addresses a different source of ambiguity, as shown in Figure~\ref{fig:common-vs-global}(c). At the reference point, consider:
$$
D_t^\star
=
\begin{pmatrix}
0 & 1 & 1\\
0 & 1 & 2\\
1 & 0 & 0
\end{pmatrix}.
$$
The true coordinates $Z_{t,1}$ and $Z_{t,2}$ have the same structural signature $((1,0),1,0)$, while $Z_{t,3}$ has signature $((0,1),0,1)$. In the learned representation, $\widehat Z_{t,2}$ and $\widehat Z_{t,3}$ share the first signature, and $\widehat Z_{t,1}$ has the second. The nonzero pattern of $D_t^\star$ supports both:
$$
\sigma_t:
\quad
Z_{t,1}\mapsto\widehat Z_{t,2},
\quad
Z_{t,2}\mapsto\widehat Z_{t,3},
\quad
Z_{t,3}\mapsto\widehat Z_{t,1},
$$
$$
\widetilde{\sigma}_t:
\quad
Z_{t,1}\mapsto\widehat Z_{t,3},
\quad
Z_{t,2}\mapsto\widehat Z_{t,2},
\quad
Z_{t,3}\mapsto\widehat Z_{t,1}.
$$
Both permutations preserve the group, incoming, and outgoing roles. Their difference only exchanges coordinates with the same structural signature. Thus, Lemma~\ref{lem:common-support-alignment} requires a common permutation for the three structural views and does not require this permutation to be unique.

This distinction also clarifies the role of path connectedness in related temporal identification results. \citep{li2025identification2} uses transition cross Hessians and structural sparsity to identify a temporal
active support and includes path connectedness among its assumptions. Its active support combines observation and temporal participation. Our analysis keeps $\mathbb A$, $\mathbb B$, and $\mathbb C$ as separate structural
objects and then aligns them through Lemma~\ref{lem:common-support-alignment}. The resulting guarantee is sufficient for identifying the structural roles studied here, while global component-wise disentanglement requires a stronger form of permutation consistency.

\paragraph{Temporal coordinate consistency.}
The global permutation discussion above concerns consistency over the latent domain at a fixed time point. A stronger temporal requirement would use one permutation $\sigma$ over the full sequence, so that $\sigma_t=\sigma$ for all $t$. Our main results allow $\sigma_t$ to depend on $t$: Lemma~\ref{lem:common-support-alignment} aligns $\mathbb A_t$, $\mathbb B_t$, and $\mathbb C_t$ under the same permutation at each time point, while the argument is applied separately to each $D_t=J_{h_t^{-1}}$. The relation $\widehat M_t=D_{t+1}^{\top}M_tD_t$ is also compatible with distinct coordinate relabelings at adjacent times, since its rows and columns follow the coordinates at $t+1$ and $t$. The present guarantee therefore identifies the structural role of each latent coordinate at every time point. Tracking a persistent latent identity over the full sequence requires additional conditions that link the reparameterizations across time, such as a shared latent reparameterization or suitable continuity conditions over a common temporal context.

\section{Extensions of the Temporal Model}
\label{app:extensions}

The main analysis uses a first-order delayed process to isolate the core
structural argument.  This section describes how the same objects can be
adapted to richer temporal models.

\subsection{Higher-order temporal dependencies}
\label{app:higher-lags}
Higher-order dynamics allow latent coordinates to interact across several time lags. Let $L$ be a positive integer denoting the maximum lag and write $z_{a:b}:=(z_a,\ldots,z_b)$ for $a\leq b$. We retain the observation model $X_t=g_t(Z_t)$ and extend the dynamics of each coordinate $i\in[N]$ to:
\begin{equation}
Z_{t,i}=f_i\bigl(\mathrm{Pa}_d(Z_{t,i}),\epsilon_{t,i}\bigr),
\quad
\mathrm{Pa}_d(Z_{t,i})\subseteq\bigcup_{\ell=1}^{L}Z_{t-\ell},
\quad t=L+1,\ldots,T.
\label{eq:ext-higher-dgp}
\end{equation}
The initial block $Z_{1:L}$ follows a joint initial distribution, and the subsequent noises remain mutually independent and independent of this block. We assume no instantaneous dependencies in this subsection. The transition density is $p(z_t\mid z_{t-L:t-1})$. For clarity, take $T>2L$ and consider times $L+1\leq t\leq T-L$, where all required transitions are available. Related temporal identification frameworks also accommodate longer conditioning histories.

The temporal roles can summarize participation across all lags. Let $G_t^{d,[\ell]}\in\{0,1\}^{N\times N}$ satisfy $(G_t^{d,[\ell]})_{i,j}=1$ iff $Z_{t-\ell,j}\in\mathrm{Pa}_d(Z_{t,i})$. We extend the incoming and outgoing supports to:
\begin{equation}
\begin{aligned}
\mathcal S_t^{\mathbb B}
&:=\bigl\{i\in[N]:\exists\ell\in[L],j\in[N]\ \text{s.t.}\ (G_t^{d,[\ell]})_{i,j}=1\bigr\},\\
\mathcal S_t^{\mathbb C}
&:=\bigl\{i\in[N]:\exists\ell\in[L],j\in[N]\ \text{s.t.}\ (G_{t+\ell}^{d,[\ell]})_{j,i}=1\bigr\}.
\end{aligned}
\label{eq:ext-higher-supports}
\end{equation}
As before, $\mathbb B_{t,i}=\mathbf 1\{i\in\mathcal S_t^{\mathbb B}\}$ and $\mathbb C_{t,i}=\mathbf 1\{i\in\mathcal S_t^{\mathbb C}\}$. These roles record whether a coordinate has any delayed parent or child within the lag range. The group roles $\{\mathbb A_{t,i}^{(k)}\}_{k=1}^{K}$ remain unchanged.
The derivative objects must condition on the full history. For each valid transition and $\ell\in[L]$:
\begin{equation}
M_t^{[\ell]}
:=\nabla_{z_{t+\ell}}\nabla_{z_t}^{\top}
\log p\bigl(z_{t+\ell}\mid z_{t+\ell-L:t+\ell-1}\bigr).
\label{eq:ext-lag-hessian}
\end{equation}
All other history coordinates are held fixed when taking these derivatives. This conditioning is important for preserving the interpretation of a direct delayed edge. For example, a path through $Z_{t+1}$ can make $Z_{t+2}$ depend on $Z_t$ after $Z_{t+1}$ is marginalized out, even when the graph has no direct edge from $Z_t$ to $Z_{t+2}$. We therefore use the conditional density in Eq.~\ref{eq:ext-lag-hessian} to characterize influence at lag $\ell$. The learned blocks $\widehat M_t^{[\ell]}$ are defined analogously.

The same change of latent coordinates applies to every lag. Suppose that the true and learned models are observationally equivalent, their transition densities are positive and $C^2$, and their observation embeddings induce the $C^1$ diffeomorphisms $h_s=\widehat g_s^{-1}\circ g_s$. Write $D_s:=J_{h_s^{-1}}(\widehat z_s)$. The conditional density transformation gives:
\begin{equation}
\widehat M_t^{[\ell]} = D_{t+\ell}^{\top}M_t^{[\ell]}D_t.
\label{eq:ext-lag-transform}
\end{equation}
The log Jacobian term depends only on the target state and disappears after differentiation with respect to the history coordinate. Thus, the calculation in Appendix~\ref{app:proof-temporal} extends to every lag.

The incoming and outgoing blocks can be collected around the same time point. Suppressing their state arguments, define:
\begin{equation}
\mathcal M_t^{\mathbb B}
:=\begin{bmatrix}
M_{t-1}^{[1]}&\cdots&M_{t-L}^{[L]}
\end{bmatrix},
\quad
\mathcal M_t^{\mathbb C}
:=\begin{bmatrix}
(M_t^{[1]})^{\top}&\cdots&(M_t^{[L]})^{\top}
\end{bmatrix}^{\top}.
\label{eq:ext-block-hessians}
\end{equation}
These matrices have dimensions $N\times LN$ and $LN\times N$. Assume the counterpart of cross Hessian faithfulness for each lag, $\operatorname{supp}(M_t^{[\ell]})=\operatorname{supp}(G_{t+\ell}^{d,[\ell]})$, in both models. The row support of $\mathcal M_t^{\mathbb B}$ and the column support of $\mathcal M_t^{\mathbb C}$ then coincide with the corresponding temporal role supports. Let $\Delta_t^-:=\operatorname{diag}(D_{t-1},\ldots,D_{t-L})$ and $\Delta_t^+:=\operatorname{diag}(D_{t+1},\ldots,D_{t+L})$. Eq.~\ref{eq:ext-lag-transform} yields:
\begin{equation}
\widehat{\mathcal M}_t^{\mathbb B}
=D_t^{\top}\mathcal M_t^{\mathbb B}\Delta_t^-,
\quad
\widehat{\mathcal M}_t^{\mathbb C}
=(\Delta_t^+)^{\top}\mathcal M_t^{\mathbb C}D_t.
\label{eq:ext-block-transform}
\end{equation}
Both temporal views therefore constrain the same matrix $D_t$.
A sufficient variation condition can be stated directly for the role directions. For every fixed $z_t^\star\in\mathcal Z_t$, suppose that:
\begin{equation}
\begin{aligned}
\operatorname{span}\bigl\{ [\mathcal M_t^{\mathbb B}]_{:,j}:
j\in[LN],\ z_t=z_t^\star \bigr\}
&=\mathbb R^N_{\mathcal S_t^{\mathbb B}},\\
\operatorname{span}\bigl\{ [\mathcal M_t^{\mathbb C}]_{j,:}^{\top}: j\in[LN],\ z_t=z_t^\star \bigr\}
&=\mathbb R^N_{\mathcal S_t^{\mathbb C}}.
\end{aligned}
\label{eq:ext-role-variation}
\end{equation}
Here, all state arguments other than $z_t$ range over the admissible support. The first condition separates the incoming directions using the history, while the second separates the outgoing directions using the remaining temporal context. When $L=1$, the matrix spanning assumptions in Theorem~\ref{thm:outgoing-temporal-support} and Corollary~\ref{cor:incoming-temporal-support} imply these conditions.

These conditions recover the support separation used in the main analysis. If $v\notin\widehat{\mathcal S}_t^{\mathbb B}$, the $v$th row of $\widehat{\mathcal M}_t^{\mathbb B}$ vanishes throughout its domain. Since $\Delta_t^-$ is nonsingular, Eq.~\ref{eq:ext-block-transform} gives $(D_te_v)^{\top}\mathcal M_t^{\mathbb B}=0$. Similarly, $v\notin\widehat{\mathcal S}_t^{\mathbb C}$ implies $\mathcal M_t^{\mathbb C}D_te_v=0$. At fixed $z_t$, $D_t$ is fixed, so Eq.~\ref{eq:ext-role-variation} forces $(D_t)_{i,v}=0$ for the true active indices $i$. With $|\widehat{\mathcal S}_t^{\mathbb B}|\leq|\mathcal S_t^{\mathbb B}|$ and $|\widehat{\mathcal S}_t^{\mathbb C}|\leq|\mathcal S_t^{\mathbb C}|$, the supported permutation argument applies. Together with the assumptions of Theorem~\ref{thm:group-support}, it gives one permutation $\sigma_t$ s.t.
\begin{equation}
\widehat{\mathcal S}_{t,k}^{\mathbb A}
=\sigma_t(\mathcal S_{t,k}^{\mathbb A}),
\quad
\widehat{\mathcal S}_t^{\mathbb B}
=\sigma_t(\mathcal S_t^{\mathbb B}),
\quad
\widehat{\mathcal S}_t^{\mathbb C}
=\sigma_t(\mathcal S_t^{\mathbb C}).
\label{eq:ext-common-alignment}
\end{equation}
Consequently, the joint signature and its role classes are preserved under the same relabeling at time $t$.
This extension retains the original granularity of the temporal roles. Identifying the incoming and outgoing supports separately for each lag would require variation and minimality conditions for those separate supports. For estimation, an extension of \textsc{RoleTrace} can condition its transition model on $L$ previous states and compute role scores from the blocks in Eq.~\ref{eq:ext-block-hessians}, using one common coordinate transformation for all blocks involving $Z_t$.

\subsection{Instantaneous dependencies}
\label{app:instantaneous}
Instantaneous dependencies allow latent coordinates to influence one another within the same time step. We return to a temporal lag of one and consider:
\begin{equation}
Z_{t,i}
=f_i\bigl(\mathrm{Pa}_d(Z_{t,i}),\mathrm{Pa}_e(Z_{t,i}),\epsilon_{t,i}\bigr),
\quad
\mathrm{Pa}_d(Z_{t,i})\subseteq Z_{t-1},
\label{eq:ext-instant-dgp}
\end{equation}
for $t=2,\ldots,T$ and $i\in[N]$, where $\mathrm{Pa}_e(Z_{t,i})\subseteq Z_t\setminus\{Z_{t,i}\}$ denotes the instantaneous parents. We assume that these edges form a directed acyclic graph within each time step and retain mutually independent innovations that are independent of the past. The observation model remains $X_t=g_t(Z_t)$. The supports $\mathcal S_t^{\mathbb B}$ and $\mathcal S_t^{\mathbb C}$ again refer to direct delayed edges between adjacent times.

The observation argument and the temporal transformation rule remain available. Under the assumptions of Theorem~\ref{thm:group-support}, group support identification follows from the same decoder Jacobian argument. Positive $C^2$ transition densities and the induced diffeomorphisms also give $\widehat M_{t-1}=D_t^{\top}M_{t-1}D_{t-1}$. This identity follows the change-of-variables formula for the joint conditional density and does not require the coordinates of $Z_t$ to be conditionally independent given $Z_{t-1}$.

The interpretation of incoming support requires additional effort. Let $P_t^e(i):=\{j\in[N]:Z_{t,j}\in\mathrm{Pa}_e(Z_{t,i})\}$. The transition density factorizes as:
\begin{equation}
p(z_t\mid z_{t-1}) =\prod_{i=1}^{N} p\bigl(z_{t,i}\mid z_{t,P_t^e(i)},z_{t-1}\bigr),
\label{eq:ext-instant-factorization}
\end{equation}
where each factor depends on the past only through its delayed parents. An instantaneous parent appears in the factor of its child, so differentiating this factor can create a nonzero incoming row for a coordinate without a delayed parent. For example, consider:
\begin{equation}
Z_{t,1}=\epsilon_{t,1}, \quad
Z_{t,2}=aZ_{t-1,1}+bZ_{t,1}+\epsilon_{t,2},
\quad ab\neq0,
\label{eq:ext-instant-example}
\end{equation}
with independent standard Gaussian noises. Here, $(\mathbb B_{t,1},\mathbb B_{t,2})=(0,1)$, while direct differentiation of the conditional density gives:
\begin{equation}
[M_{t-1}]_{:,1}
=\begin{pmatrix}-ab\\a\end{pmatrix}.
\label{eq:ext-instant-example-hessian}
\end{equation}
Both rows are nonzero. The additional row arises because $Z_{t,1}$ and $Z_{t-1,1}$ enter the same conditional mechanism for $Z_{t,2}$. Thus, the row support can differ from the set of direct delayed receivers.
Therefore, a restriction on instantaneous parents preserves the zero rows needed by our argument. Specifically, suppose that, at each relevant time, both models satisfy:
\begin{equation}
P_t^e(i)\subseteq\mathcal S_t^{\mathbb B}
\quad\text{for every }i\in\mathcal S_t^{\mathbb B},
\label{eq:ext-instant-parent-condition}
\end{equation}
with the corresponding hatted quantities in the learned model. This condition requires every instantaneous parent of a delayed receiver to have a delayed parent as well. It allows instantaneous edges among delayed receivers and allows them to influence other coordinates within the same time step. Under Eq.~\ref{eq:ext-instant-parent-condition}, every factor in Eq.~\ref{eq:ext-instant-factorization} that depends on the past involves current coordinates only from $\mathcal S_t^{\mathbb B}$. Hence,
$i\notin\mathcal S_t^{\mathbb B}
\;\Longrightarrow\;
[M_{t-1}]_{i,:}\equiv0.$
The same implication holds in the learned model. On the outgoing side, a coordinate without any delayed child appears in none of the conditional factors for the next state, which already gives $i\notin\mathcal S_t^{\mathbb C}\Rightarrow[M_t]_{:,i}\equiv0$.

Joint role identification then follows under variation and minimality at the level of these supports. In addition to the smoothness and observation assumptions above, impose Eq.~\ref{eq:ext-role-variation} with $L=1$, $\mathcal M_t^{\mathbb B}=M_{t-1}$, and $\mathcal M_t^{\mathbb C}=M_t$, together with $|\widehat{\mathcal S}_t^{\mathbb B}|\leq|\mathcal S_t^{\mathbb B}|$ and $|\widehat{\mathcal S}_t^{\mathbb C}|\leq|\mathcal S_t^{\mathbb C}|$. The zero rows and columns established above yield:
$$
(D_te_v)^{\top}M_{t-1}=0
\quad\text{if }v\notin\widehat{\mathcal S}_t^{\mathbb B},
\quad
M_tD_te_v=0
\quad\text{if }v\notin\widehat{\mathcal S}_t^{\mathbb C}.
$$
The spanning conditions again force the corresponding entries of $D_t$ to vanish. Evaluating at the observation reference point and using the supported permutation argument in the proof of Lemma~\ref{lem:common-support-alignment} gives the common alignment in Eq.~\ref{eq:ext-common-alignment}. 

More general instantaneous models can be approached through richer causal identification conditions. Existing work uses temporal context, sufficient variability, and sparse influence constraints to identify latent components in the presence of instantaneous dependence~\citep{li2025identification}. Under their respective assumptions, these results provide a complementary route in which temporal roles can be read from a recovered delayed graph. The additional restriction in Eq.~\ref{eq:ext-instant-parent-condition} specifies the class considered by the support argument above.
\section{Related Work}
\label{sec:related-work}
In this section, we review identification results for latent variable models, temporal processes, and partially observed systems, and explain how they relate to the structural roles studied in this work.

\paragraph{Latent Variable Identification and Observation Structure.}
Latent variable identification studies which properties of a hidden generating process can be recovered from observations. Independent component analysis (ICA) provides a classical formulation, in which statistically independent sources are recovered from their linear mixtures under suitable distributional conditions~\citep{COMON1994287,10.1162/153244303768966085,d243f4c2f46949b29cf034348d84c11f}. With general nonlinear mixing, independence alone leaves substantial ambiguity in the recovered sources~\citep{HYVARINEN1999429}. Auxiliary variable approaches obtain identification from conditional independence and sufficient variation across an observed variable~\citep{Hyvrinen2018NonlinearIU,pmlr-v108-khemakhem20a}, while other results restrict the generating function~\citep{BucBesSch22,lachapelle2023additive}. Structural assumptions on the observation map provide another route. 
Structural sparsity constrains the dependency pattern between sources and observed variables, enabling identification under suitable conditions on the support of the mixing Jacobian~\cite{li2025identification,3666122.3666707}. G-CaRL uses a known grouping of observational variables to identify latent causal variables and their relations, with observed and latent variables partitioned into corresponding groups~\citep{pmlr-v235-morioka24a}. Multiple views provide another source of identification information~\citep{GreRubMehLocSch19,3692070.3694554}, while~\citep{zheng2025thought} identifies shared and private latent information together with the incidence structure between latent factors and observation groups. 
Our setting allows one latent coordinate to be expressed in several groups and retains its complete incidence profile $\{\mathbb A_{t,i}^{(k)}\}_{k=1}^K$. We study this observation structure jointly with incoming and outgoing temporal participation, preserving the correspondence among all three attributes.

\paragraph{Temporal Structure for Latent Identification.}
Temporal observations provide identification information through changes in source distributions and dependence across time. One line of work exploits observed segments, latent regimes, or multiple distributions to identify nonlinear representations from nonstationary data~\citep{3157382.3157519, yao2022learning, 3666122.3666477,pmlr-v323-brouillard26a}. Others use dependence across consecutive states or independent innovations as the identifying signal, covering stationary nonlinear ICA and more general autoregressive processes~\citep{pmlr-v54-hyvarinen17a,pmlr-v130-morioka21a,3600270.3602191}. Historical context can recover latent states when a single observation loses information~\citep{3692070.3692351}. Sparse temporal mechanisms support identification of latent components and dependency structure~\citep{pmlr-v177-lachapelle22a,li2025identification,JMLR:v27:24-0771}. Intervention-based approaches provide another source of temporal structure and establish identification of causal factors and their instantaneous or delayed relations~\citep{pmlr-v162-lippe22a,lippe2023causal}. 
These methods characterize latent components, mechanisms, or causal edges at different levels of granularity.
Our target is complementary: we characterize each latent coordinate by whether it receives delayed influence and whether it transmits delayed influence, represented by $\mathbb B_{t,i}$ and $\mathbb C_{t,i}$.
This factorized temporal role preserves distinctions between coordinates that receive, transmit, or locally terminate temporal information, and it can be combined with observation structure without requiring recovery of every delayed causal edge.

\paragraph{Partial Observation and Latent Participation.}
Several recent formulations allow latent information to be exposed only through selected views, instances, or time steps. 
MuLTI studies overlapping latent processes across multiple time series views and learns permutation-based correspondence between view specific estimates~\citep{10.24963/ijcai.2023/428}. Multi-view causal representation learning identifies information shared across subsets of simultaneously observed views under partial observability~\citep{yao2024multiview}, while sparsity based approaches allow the visible subset to vary across unpaired observations~\citep{pmlr-v235-xu24ac}. InterLatent considers a time varying active support defined jointly through participation in the observation and adjacent transition mechanisms~\citep{li2025identification2}. These works target view correspondence, partially visible latent subsets, or a single active support. 
Our formulation keeps group expression $\mathbb A$, incoming influence $\mathbb B$, and outgoing influence $\mathbb C$ as separate structural objects and identifies their correspondence under one common permutation. This separation distinguishes latent coordinates with the same observation profile but different temporal roles, including factors that only receive or only transmit delayed influence.

\section{Additional Experimental Details}
\label{app:exp}

\subsection{Additional Details for Synthetic Experiments}
\label{app:synthetic-setup}
This section specifies the synthetic mechanisms, estimator, and evaluation, followed by additional results for RQ1 and the controlled designs for RQ2 and RQ3.

\paragraph{Synthetic Data Generation.}
The RQ1 datasets are generated from independently sampled random mechanisms. Within each mechanism, the observation map and transition law are fixed across time, so its group structure and delayed graph remain constant, and we suppress the time index on the roles below. This construction is time homogeneous; the initial distribution need not be the stationary distribution. Figure~\ref{fig:exp-rq1-dgp} illustrates one sampled mechanism. RQ2 uses different graphs on its two transitions, as specified separately below.

We set $N=6$, $K=3$, and $T=5$, with six observed dimensions in each group. The initial latent state is sampled as $Z_1\sim\mathcal N(0,I_N)$. Let $\mathbf A\in\{0,1\}^{K\times N}$ denote the group--latent incidence matrix. Each entry is sampled with probability $0.55$, subject to the conditions that every observation group contains at least one active latent coordinate and every latent coordinate is expressed in at least one group. For each $k\in[K]$, let $W^{(k)}\in\mathbb R^{6\times N}$. When $\mathbf A_{k,i}=1$, the corresponding column of $W^{(k)}$ is sampled from a scaled orthonormal basis. Columns associated with $\mathbf A_{k,i}=0$ are set to zero. The observation mechanism is:
\begin{equation}
    a_t^{(k)} = W^{(k)}Z_t,
    \quad
    X_t^{(k)} = a_t^{(k)} + 0.12\tanh\!\left(a_t^{(k)}\right).
\label{eq:app-synthetic-observation}
\end{equation}
Let $G^d\in\{0,1\}^{N\times N}$ denote the delayed graph. For every mechanism, three latent coordinates are sampled as candidate senders and three as candidate receivers. Additional admissible edges are sampled independently with probability $0.35$. The matrices $W^\mu,W^\sigma\in\mathbb R^{N\times N}$ share the support $G^d$. For every active entry, the sign is sampled independently and the magnitude is sampled uniformly from $[0.45,0.90]$. For $t\geq2$, define:
\begin{align}
\mu_i(Z_{t-1}) &= 0.75\tanh\!\left( \sum_{j=1}^{N} W^\mu_{ij} \left[ Z_{t-1,j} + 0.15\sin(Z_{t-1,j}) \right] \right),
\label{eq:app-synthetic-mu}
\\
\log\sigma_i(Z_{t-1}) &= -0.65 + 0.30\tanh\!\left( \sum_{j=1}^{N} W^\sigma_{ij} \left[ Z_{t-1,j} + 0.10\tanh(2Z_{t-1,j}) \right] \right),
\label{eq:app-synthetic-sigma}
\\
Z_{t,i} &= \mu_i(Z_{t-1}) + \sigma_i(Z_{t-1})\epsilon_{t,i}, \quad \epsilon_{t,i} \overset{\mathrm{i.i.d.}}{\sim} \mathcal N(0,1).
\end{align}
The state-dependent mean and variance make the transition cross-Hessian vary with the latent state. The temporal roles are:
\begin{equation}
    \mathbb B_i = \mathbf 1 \left\{ \sum_{j=1}^{N} G^d_{ij} > 0 \right\},
    \quad
    \mathbb C_i = \mathbf 1 \left\{ \sum_{j=1}^{N} G^d_{ji} > 0 \right\}.
\label{eq:app-role-BC}
\end{equation}
Thus, $\mathbb B_i=1$ indicates that coordinate $i$ receives delayed influence, and $\mathbb C_i=1$ indicates that coordinate $i$ transmits delayed influence. We retain 16 independently sampled mechanisms. For every mechanism, we generate one master training set and use nested prefixes of sizes $n \in \{50,100,150,200,400,800,1600,3200\}$. The validation and test sets contain 512 and 1024 additional sequences, respectively. They remain fixed across sample sizes and methods.

\paragraph{Estimator and Optimization.}
The representation model follows the procedure described in Section~\ref{sec:estimation}. The transition density is parameterized as a conditional diagonal Gaussian. Its mean and log variance networks each use two $\tanh$ hidden layers of width 32. The representation and transition stages are trained for 150 epochs using AdamW with batch size 32, initial learning rate $10^{-3}$, and weight decay $10^{-5}$. After epoch 15, a cosine schedule decreases the learning rate to $0.05$ of its initial value. Before aggregation, derivative magnitudes are normalized by the empirical standard deviation of the corresponding latent coordinates. For the common coordinate objective, we use:
\begin{equation}
    \left( \lambda_{\mathbb A}, \lambda_{\mathbb B}, \lambda_{\mathbb C} \right) = (0.004,0.0015,0.0015).
\label{eq:app-structural-weights}
\end{equation}
The common coordinate transformation is optimized using the Jacobi procedure described in Section~\ref{sec:estimation}.
After the structural scores are obtained, each score vector is transformed to log space and partitioned by two means clustering. The procedure is applied separately to each $\mathbb A$ group and separately to $\mathbb B$ and $\mathbb C$. The resulting cluster assignment determines the estimated binary support.

For \textsc{RoleTrace}, the first common coordinate update is followed by one transition refit in the updated latent coordinates. We then recompute the temporal derivatives and perform one final Jacobi update. \textsc{No-refit} uses the first common coordinate update as its final estimate. The remaining baselines isolate different sources of structural information. \textsc{No Structural} retains the initial representation and transition, while \textsc{Temporal-only} uses $\lambda_{\mathbb A}=0$ for coordinate selection. The \textsc{ThoughtComm} adaptation selects coordinates using the observation term alone and evaluates the temporal roles in those same coordinates. The \textsc{InterLatent} adaptation trains reconstruction, transition likelihood, and the three role penalties jointly in the grouped observation setting.

\paragraph{Evaluation and Additional RQ1 Results.}
\label{app:synthetic-rq1}
Evaluation compares the complete roles under a single relabeling of the learned coordinates. Let $\widehat r_i$ and $r_i$ be the estimated and true $(K+2)$ bit signatures. Hungarian matching finds:
\begin{equation}
 \sigma^*\in\arg\min_{\sigma\in\mathfrak S_N}
 \sum_{i=1}^{N}\|\widehat r_{\sigma(i)}-r_i\|_1.
 \label{eq:app-common-matching}
\end{equation}
This permutation only relabels an already fitted model for evaluation. The orthogonal coordinate transformation $Q_t$ is learned from the structural objective without ground truth, and all RQ1 metrics use the same $\sigma^*$ for the three roles.
The metrics distinguish recovery of individual supports from recovery of the complete signature. For each role, $\mathrm{F1}=2\mathrm{TP}/(2\mathrm{TP}+\mathrm{FP}+\mathrm{FN})$, with the observation entries pooled across groups. For aligned signatures, we define:
\begin{equation}
 \mathrm{Joint}=\frac1N\sum_{i=1}^{N}\mathbf1\{\widehat r_{\sigma^*(i)}=r_i\},\qquad
 \mathrm{Hamming}=1-\frac{\sum_i\|\widehat r_{\sigma^*(i)}-r_i\|_1}{N(K+2)}.
 \label{eq:app-joint-hamming}
\end{equation}
Joint accuracy counts coordinates whose entire signature is correct, while Hamming accuracy counts correct signature entries. When reported, Common F1 is the arithmetic mean of the three role F1 values under the same alignment. AUPRC uses continuous scores and is computed as average precision, $\sum_\ell (R_\ell-R_{\ell-1})P_\ell$, grouping equal scores at each threshold. Means and standard errors use mechanisms as the independent units, with $\mathrm{SE}=s/\sqrt m$ for $m$ mechanisms.

\begin{figure}[htbp]
\vspace*{-.1in}
\centering

\begin{subfigure}{0.24\textwidth}
    \centering
    \includegraphics[width=\linewidth]{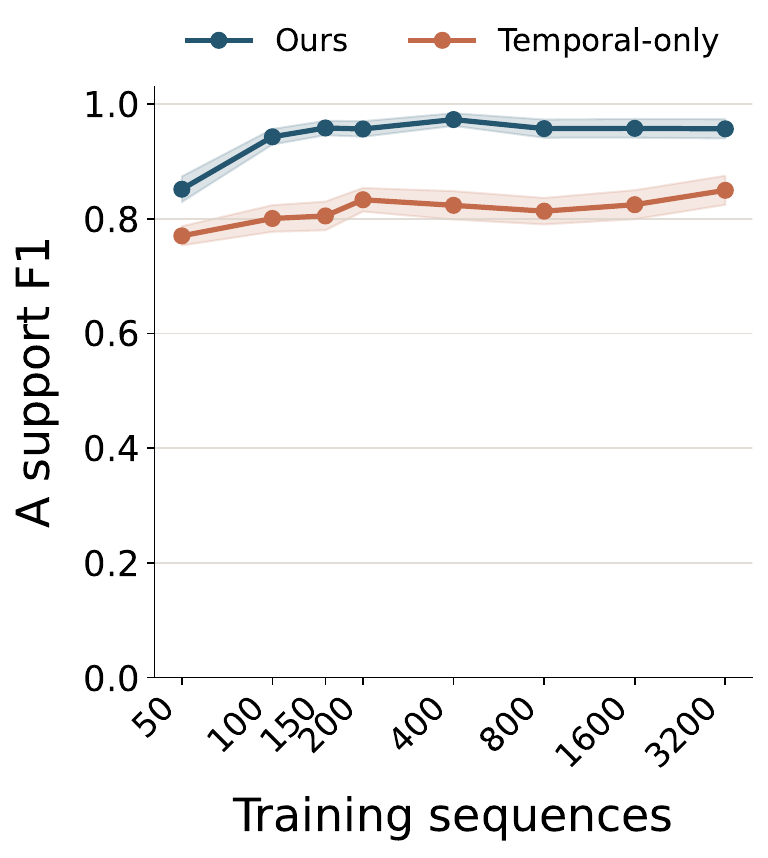}
\end{subfigure}
\begin{subfigure}{0.24\textwidth}
    \centering
    \includegraphics[width=\linewidth]{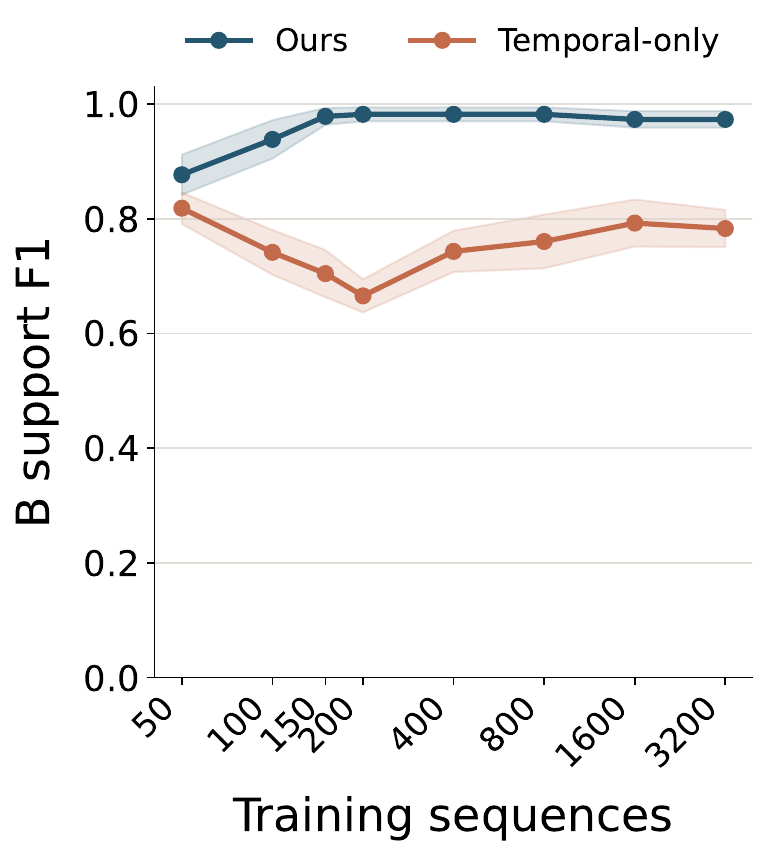}
\end{subfigure}
\begin{subfigure}{0.24\textwidth}
    \centering
    \includegraphics[width=\linewidth]{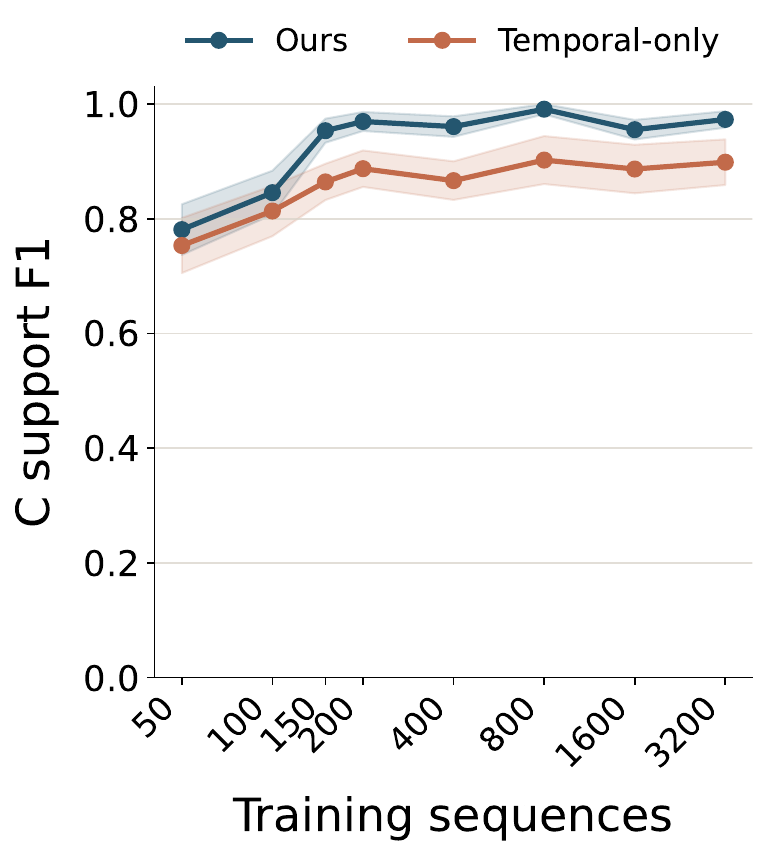}
\end{subfigure}
\begin{subfigure}{0.24\textwidth}
    \centering
    \includegraphics[width=\linewidth]{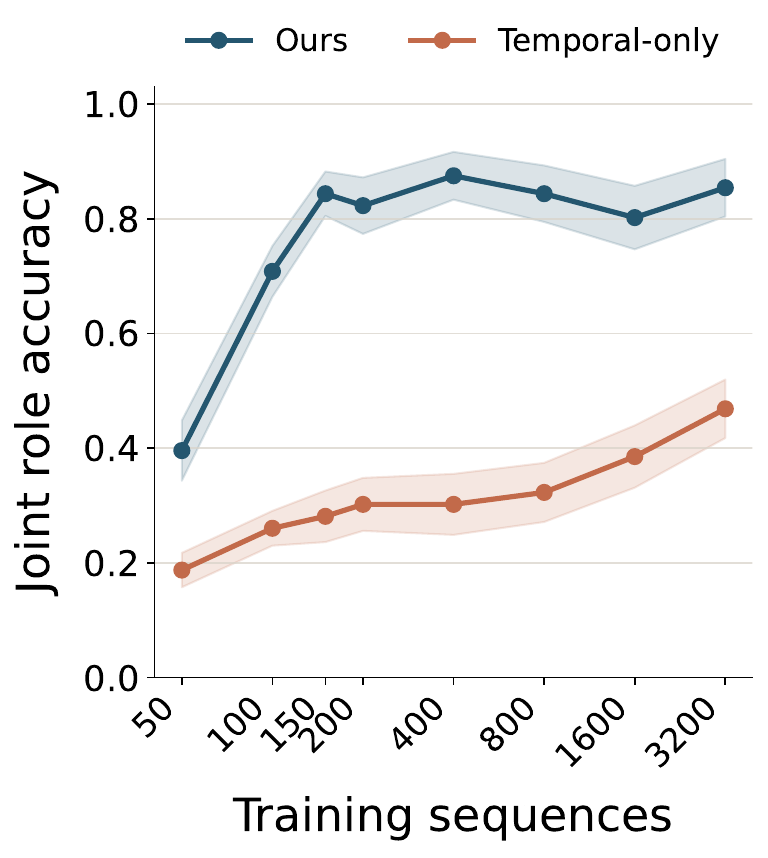}
\end{subfigure}

\caption{Sample-size scaling of structural recovery.}
\label{fig:rq1-sample-size-scaling}
\end{figure}

The sample size study shows that recovery improves most clearly in the small data regime. In Figure~\ref{fig:rq1-sample-size-scaling}, \textsc{Ours} increases from $0.396$ Joint accuracy at $n=50$ to $0.708$ at $100$ and $0.844$ at $150$. For $n\geq150$, Joint accuracy remains high and ranges from $0.802$ to $0.875$ across the tested sizes. The incoming role reaches high F1 early, while the outgoing role improves from $0.781$ at $n=50$ to $0.954$ at $150$. \textsc{Ours} exceeds \textsc{Temporal-only} in Joint accuracy at every displayed size. The curves therefore show a clear gain from additional data at small sample sizes, followed by a high but fluctuating recovery level.

\begin{figure}[t]
 \centering
 \includegraphics[width=0.7\textwidth]{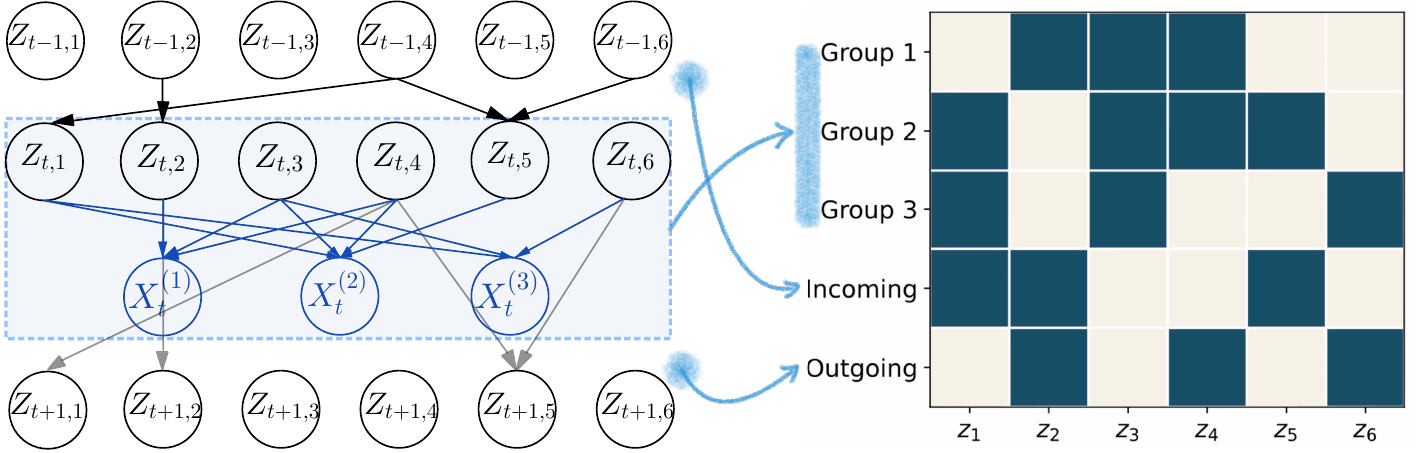}
 \caption{Data generation process and target identification of the simulation experiment.} 
 \label{fig:exp-rq1-dgp-v1}
\end{figure}

\begin{figure}[htbp]
    \centering

    \begin{subfigure}[t]{0.32\textwidth}
        \centering
        \includegraphics[width=\linewidth]{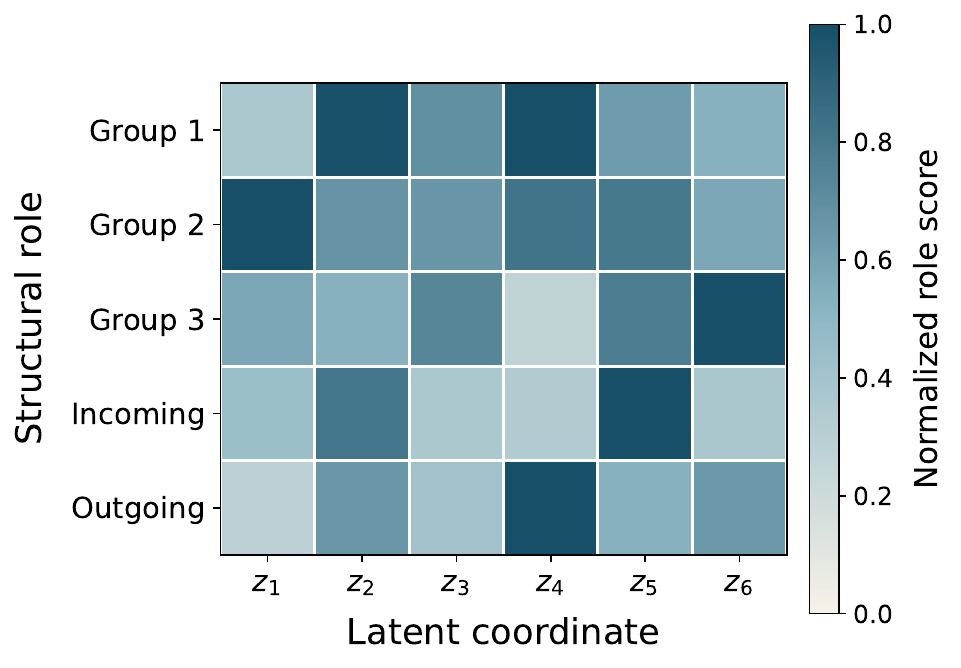}
        \caption{No structural}
    \end{subfigure}
    \hfill
    \begin{subfigure}[t]{0.32\textwidth}
        \centering
        \includegraphics[width=\linewidth]{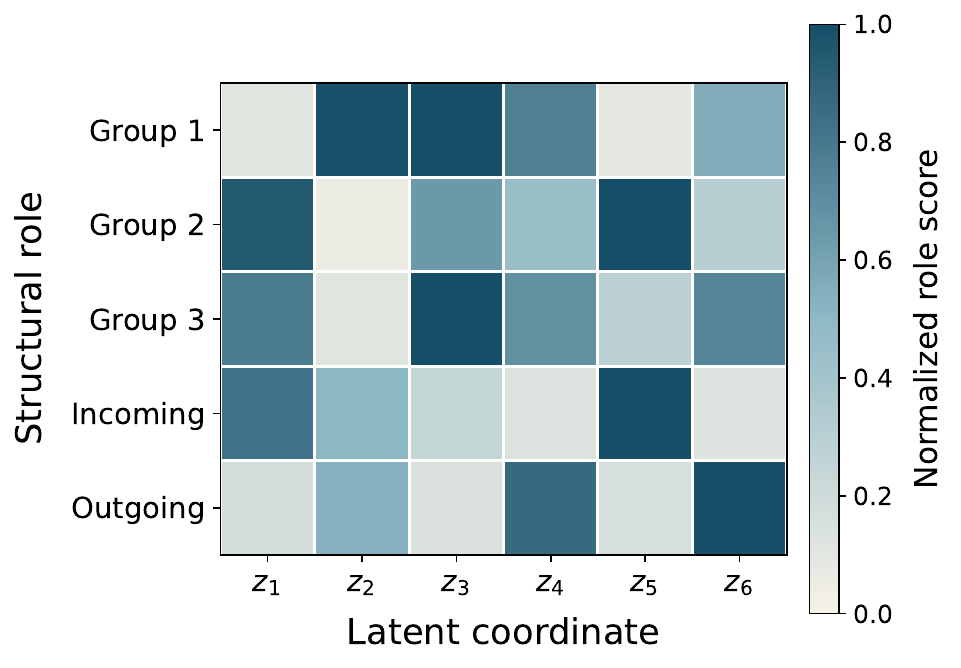}
        \caption{Temporal only}
    \end{subfigure}
    \hfill
    \begin{subfigure}[t]{0.32\textwidth}
        \centering
        \includegraphics[width=\linewidth]{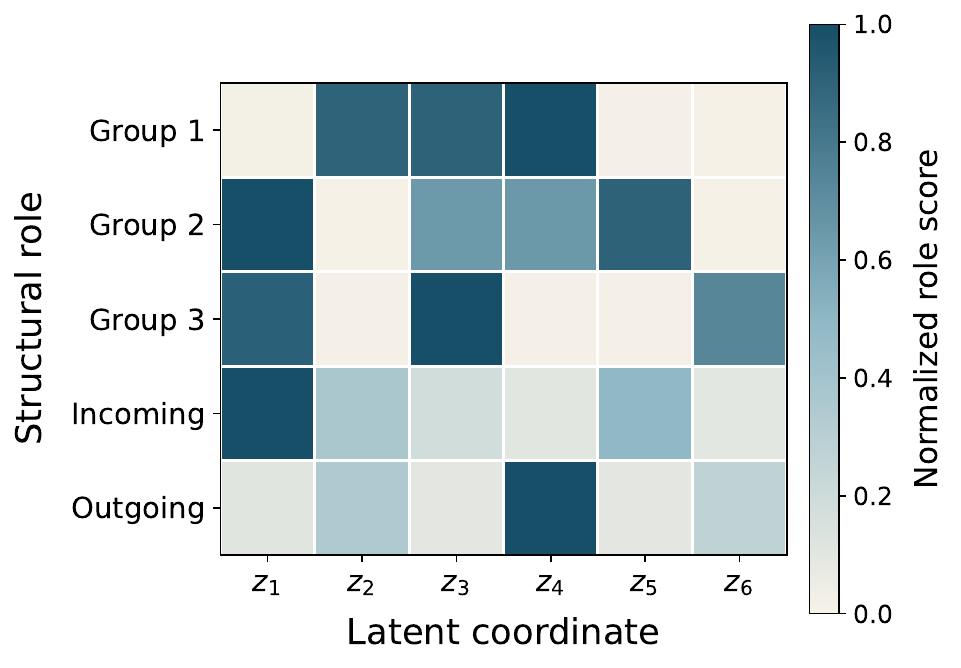}
        \caption{No refit}
    \end{subfigure}


    \begin{subfigure}[t]{0.32\textwidth}
        \centering
        \includegraphics[width=\linewidth]{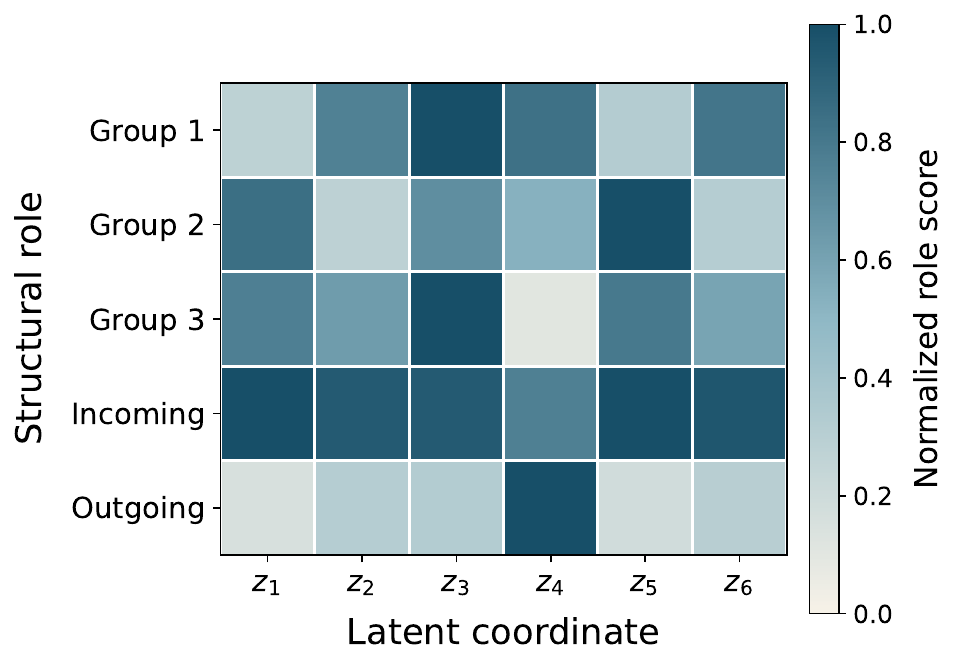}
        \caption{InterLatent}
    \end{subfigure}
    \hfill
    \begin{subfigure}[t]{0.32\textwidth}
        \centering
        \includegraphics[width=\linewidth]{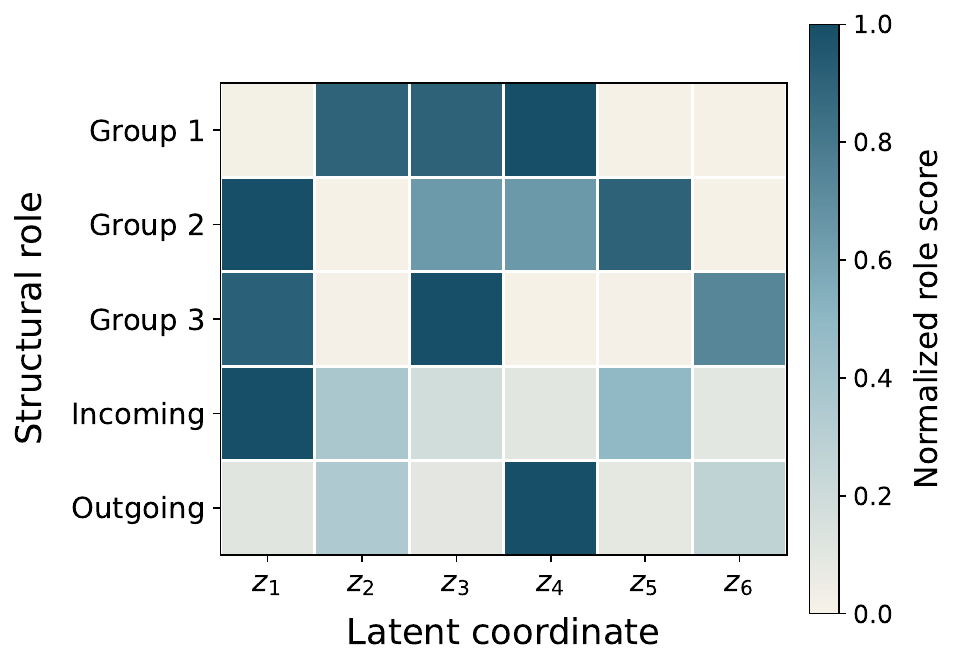}
        \caption{ThoughtComm}
    \end{subfigure}
    \hfill
    \begin{subfigure}[t]{0.32\textwidth}
        \centering
        \includegraphics[width=\linewidth]{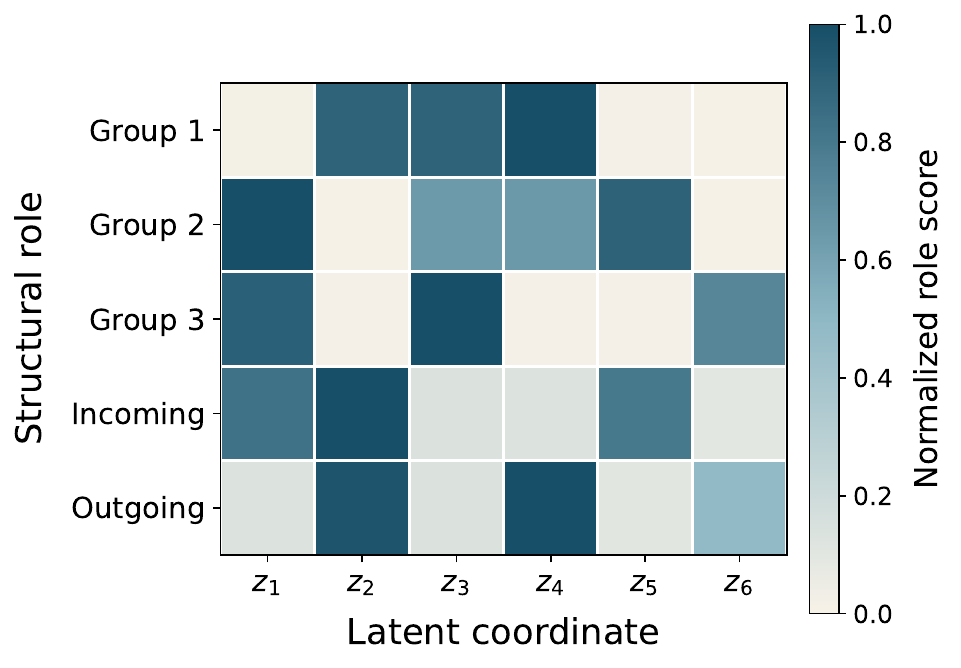}
        \caption{RoleTrace}
    \end{subfigure}

    \caption{Heatmaps of recovered structural signatures.}
    \label{fig:rq1-heatmaps-v1}
\end{figure}

Here we provide complementary examples of the continuous structural scores. Each row is divided by its maximum, so color represents relative strength within a role or observation group and should not be interpreted as a probability or binary support. Figure~\ref{fig:rq1-heatmaps-v1} shows an additional mechanism, whose generating structure appears in Figure~\ref{fig:exp-rq1-dgp-v1}.

\paragraph{Common Alignment (RQ2)}
RQ2 uses a controlled construction in which the three structural views leave different coordinate ambiguities. We use ten mechanisms with $N=4$, $K=3$, $T=3$, and four observed dimensions per group, and evaluate the middle time point $t=2$. The supports are:
\begin{equation}
 (\mathbb A^{(k)}_{2,i})_{k,i}=
 \begin{pmatrix}1&1&0&0\\0&0&1&0\\0&0&0&1\end{pmatrix},\qquad
 \mathbb B_2=(1,0,1,0),\qquad \mathbb C_2=(0,1,1,1).
\end{equation}
Using indices starting at one, the first transition has edges $Z_{1,2}\to Z_{2,1}$ and $Z_{1,4}\to Z_{2,3}$, while the second has $Z_{2,2}\to Z_{3,2}$, $Z_{2,3}\to Z_{3,3}$, and $Z_{2,4}\to Z_{3,4}$. Both transitions use the nonlinear Gaussian construction above with separately sampled weights. Training uses 3200 sequences and the same validation, test, and optimization settings as RQ1. Separate temporal networks fit the two transitions. Refining the middle coordinate system transforms the derivative objects as $J_{\widehat g_2}Q_2$, $Q_2^\top\widehat M_1$, and $\widehat M_2Q_2$. For $\mathbb X\in\{\mathbb A,\mathbb B,\mathbb C\}$, let $c_{\mathbb X}(\sigma)$ count mismatched entries in that role and define:
\begin{equation}
 \Pi_{\mathbb X}=\arg\min_{\sigma\in\mathfrak S_4}c_{\mathbb X}(\sigma),\qquad
 \Pi_{\mathrm{joint}}=\arg\min_{\sigma\in\mathfrak S_4}
 \sum_{\mathbb X\in\{\mathbb A,\mathbb B,\mathbb C\}}c_{\mathbb X}(\sigma).
 \label{eq:app-rq2-optimal-permutations}
\end{equation}
We enumerate all 24 permutations and average each metric over the corresponding optimal set. The $\mathbb A$ guided, $\mathbb B$ guided, and $\mathbb C$ guided conditions use $\Pi_{\mathbb A}$, $\Pi_{\mathbb B}$, and $\Pi_{\mathbb C}$, while common alignment uses $\Pi_{\mathrm{joint}}$. \textsc{No Structural} evaluates the initial scores with common matching. The results show how a single structural view can leave the complete coordinate correspondence unresolved. With learned scores, common alignment gives mean role F1 values $(1.000,1.000,0.960)$ and Joint accuracy $0.950\pm0.033$. The respective single-role alignments give Joint accuracies of $0.700\pm0.033$, $0.475\pm0.017$, and $0.458\pm0.028$, while \textsc{No Structural} gives $0.100$.

\paragraph{Common Alignment under Different Structural Views.}
To complement the permutation analysis from the perspective of the learned coordinate transformation, we compare which structural views are used to refine the common coordinate system $Q_t$. Starting from the same representation and derivatives, we optimize $Q_t$ using $\mathbb A$, $\mathbb B$, $\mathbb C$, $\{\mathbb B,\mathbb C\}$, or all three views, while keeping $Q_t=I$ as a reference. In particular, the full objective takes the form:
$$\Omega_t(Q_t) = \lambda_A\sum_{k,i}s^A_{t,k,i}(Q_t)+\lambda_B\sum_i s^B_{t,i}(Q_t)+\lambda_C\sum_i s^C_{t,i}(Q_t),$$
with the corresponding terms removed for the partial objectives. Transition refitting is disabled so that the comparison isolates the effect of coordinate refinement.

\begin{table}[t]
\centering
\small
\setlength{\tabcolsep}{3pt}
\caption{Coordinate refinement with fixed representations and pre-refit derivatives. Values are mean $\pm$ SE across ten mechanisms. Bold denotes the largest mean in each column, including ties.}
\label{tab:rq2-coordinate-refinement}
\begin{tabular}{lccccc}
\toprule
Refinement & A F1 & B F1 & C F1 & Joint Acc. & Hamming \\
\midrule
None & $0.694 \pm 0.025$ & $0.777 \pm 0.070$ & $0.770 \pm 0.060$ & $0.100 \pm 0.055$ & $0.725 \pm 0.019$ \\
$\mathbb A$ only & $\mathbf{1.000 \pm 0.000}$ & $\mathbf{0.967 \pm 0.033}$ & $0.910 \pm 0.053$ & $0.875 \pm 0.056$ & $0.975 \pm 0.011$ \\
$\mathbb B$ only & $0.792 \pm 0.029$ & $0.867 \pm 0.069$ & $0.920 \pm 0.033$ & $0.450 \pm 0.082$ & $0.850 \pm 0.032$ \\
$\mathbb C$ only & $0.822 \pm 0.022$ & $0.947 \pm 0.037$ & $\mathbf{0.960 \pm 0.027}$ & $0.575 \pm 0.065$ & $0.890 \pm 0.024$ \\
RoleTrace & $\mathbf{1.000 \pm 0.000}$ & $\mathbf{0.967 \pm 0.033}$ & $0.930 \pm 0.052$ & $\mathbf{0.900 \pm 0.055}$ & $\mathbf{0.980 \pm 0.011}$ \\
\bottomrule
\end{tabular}
\end{table}

Table~\ref{tab:rq2-coordinate-refinement} shows that joint refinement achieves the highest Joint accuracy of $0.900$, compared with $0.875$, $0.450$ and $0.575$ using $\mathbb A$, $\mathbb B$ or $\mathbb C$ alone. These results indicate that combining observation and temporal structure can further resolve the coordinate correspondence, with the observation structure providing a particularly informative constraint in this controlled setting.

\paragraph{Sensitivity to Identifiability Conditions (RQ3).}
RQ3 isolates the structural recovery step by applying the same coordinate selection procedure directly to derivatives of the generating mechanisms. We use $N=4$, $K=3$, $T=5$, four observed dimensions per group, and 512 latent sequences evaluated at $t=3$. Each of the three degradation axes contains ten mechanisms, ten levels, and three orthogonal coordinate mixings, giving 900 evaluations in total. No neural model is trained. The transition mean scale and log standard deviation amplitude are $0.55$ and $0.20$, with intercept $-0.65$.

In Figure~\ref{fig:rq3}, the first path weakens the group span condition by making two active directions in one observation group increasingly dependent. For selected original columns $u,v$, we set:
\begin{equation}
 w_p(\gamma)=u,\qquad
 w_q(\gamma)=\sqrt{1-\gamma^2}\,u+\gamma v,
\end{equation}
while leaving the remaining columns unchanged. Let $J_{k_*,\gamma}^{(s)}$ denote the affected group's Jacobian restricted to its declared active columns. The diagnostic is
\begin{equation}
 \rho_{\mathrm{group}}(\gamma)=\min_{s\leq S}
 \frac{\sigma_{\min}(J_{k_*,\gamma}^{(s)})}{\sigma_{\max}(J_{k_*,\gamma}^{(s)})}.
 \label{eq:app-rq3-group-ratio}
\end{equation}
At $\gamma=0$, those active columns lose rank while the complete observation map remains full rank, directly weakening the group variability condition in Theorem~\ref{thm:group-support}.

The second path weakens temporal variability by approaching a linear Gaussian transition. With $\ell=\log\sigma$ and $\ell_0=-0.65$, define
\begin{equation}
 \mu_\gamma(z)=(1-\gamma)W_0z+\gamma\mu_1(z),\qquad
 \ell_\gamma(z)=\ell_0+\gamma\big[\ell_1(z)-\ell_0\big].
\end{equation}
The delayed graph has edge set $E=\{(1,3),(2,3),(4,1),(4,2)\}$, where the first index denotes the receiver, and $W_0$ preserves this support. Writing $\operatorname{vec}_E$ for vectorization restricted to these edges:
\begin{align}
 (V_\gamma^+)_{s,:}&=\operatorname{vec}_{E}M_{t-1,\gamma}(z_t^{(s)},z_{t-1}^{(1)}),\\
 (V_\gamma^-)_{s,:}&=\operatorname{vec}_{E}M_{t,\gamma}(z_{t+1}^{(1)},z_t^{(s)}).
\end{align}
For $\kappa(V)=\sigma_{\min}(V)/\sigma_{\max}(V)$, the displayed ratio is:
\begin{equation}
 \rho_{\mathrm{span}}(\gamma)=
 \frac{\min\{\kappa(V_\gamma^+),\kappa(V_\gamma^-)\}}
 {\min\{\kappa(V_1^+),\kappa(V_1^-)\}}.
 \label{eq:app-rq3-span-ratio}
\end{equation}
At $\gamma=0$, the cross Hessian is constant, so its span has dimension one within the four-dimensional edge space.

The third path weakens faithfulness through exact cancellation on one declared edge. The graph contains edges $(1,2)$ and $(3,4)$, with the first affected. For each mean or log scale coefficient $w$ on this edge, two pathways have coefficients $(1+c)w$ and $-(c+1-\gamma)w$, where $c=0.35$, so their combined coefficient is $\gamma w$. The second edge remains fixed. For the selected edge $(i_*,j_*)$, define:
\begin{equation}
 e_\gamma=\left[\frac1S\sum_{s=1}^{S}\big(M_{t-1,\gamma}^{(s)}\big)_{i_*j_*}^{2}\right]^{1/2},\qquad
 \rho_{\mathrm{edge}}(\gamma)=e_\gamma/e_1.
 \label{eq:app-rq3-edge-ratio}
\end{equation}
The target graph remains fixed at $\gamma=0$, although the selected edge then has zero net influence on the transition distribution. Recovery at this endpoint is therefore evaluated against a declared role contribution that has become unobservable.

The three paths produce distinct degradation patterns. Mean Joint accuracy falls from one to $0.633$, $0.750$, and $0.500$ at the group, temporal span, and cancellation endpoints, respectively. Group degradation primarily affects $\mathbb A$, while the temporal paths affect $\mathbb B$ and $\mathbb C$. The cancellation curve remains perfect at every tested positive ratio and drops only at zero. Plateaus are expected because the target is a discrete support: derivative magnitudes may change while the binary role remains the same. Losses before the other endpoints can also reflect the fixed structural objective and support threshold. These controlled experiments characterize the sensitivity of the specified recovery procedure along the tested paths; they do not establish that each condition is necessary for every estimator.

\subsection{Additional Details for AMI Multi-party Meeting Experiment}
\label{app:ami_details}

\paragraph{Data Preprocessing and Split.}
We use the public manual annotations of the AMI Meeting Corpus~\citep{10.1007/11677482_3} and focus on scenario meetings, where four participants collaborate on a shared design project. The four functional roles, Project Manager (PM), Industrial Designer (ID), User Interface Designer (UI), and Marketing Expert (ME), define the observation groups used throughout the experiment. Speaker identifiers are mapped to these roles using the meeting metadata, so the group ordering is fixed as PM, ID, UI, and ME across meetings. Our evaluation uses 118 meetings from 30 teams. These meetings correspond to the 98 training and 20 development meetings in our original metadata split; the remaining 20 meetings from five held-out teams are not used in this experiment. The 118 meetings contain 28 teams with four project meetings and two teams with three, due to two meetings being absent from the scenario metadata. No meeting is filtered according to meeting effectiveness, speaking activity, or downstream prediction performance.

We construct five folds at the team level so that all meetings from the same team remain in the same partition. In each fold, 20 teams are used for model fitting, four for model selection, and six for evaluation. The fit partition determines all preprocessing statistics, representation parameters, and structural rankings. The selection partition is used for checkpoint selection and downstream regularization, while the evaluation partition is used only after these choices are fixed. Each team appears in the evaluation partition exactly once across the five folds. This separation prevents windows from different meetings of the same project team from appearing in both model fitting and evaluation.

We divide each meeting into non-overlapping 15-second windows, producing 14,513 windows across the 118 meetings. The 15-second interval provides a regular temporal resolution while retaining enough local conversational context for each participant. We use it as a fixed observation interval throughout the experiment. For a window ending at time $T$, the input contains valid words whose ending times fall in $(T-15,T]$, ensuring that speech extending beyond the current boundary is not included prematurely.

Each participant contributes one grouped observation. For participant $g$, the raw observation is:
\begin{equation}
X_t^{(g)}
=
\left[
u_t^{(g)},
\log(1+n_t^{(g)}),
r_t^{(g)},
q_t^{(g)}
\right]
\in \mathbb R^{35},
\label{eq:ami_group_observation}
\end{equation}
where $u_t^{(g)}\in\mathbb R^{32}$ represents the linguistic content in the current window, $n_t^{(g)}$ is the number of completed words, $r_t^{(g)}$ records the normalized time since the participant's most recent completed word, and $q_t^{(g)}$ indicates whether the participant speaks in the current window. A participant with no valid word receives a zero text representation and zero activity indicator, while the recency feature retains information about earlier speech. We keep silent windows in the sequence. Among all windows, 542 contain no active participant, while 3,033, 3,846, 4,281, and 2,811 windows contain one, two, three, and four active participants, respectively. The complete observation is therefore:
$
X_t=(X_t^{(\mathrm{PM})},X_t^{(\mathrm{ID})},
X_t^{(\mathrm{UI})},X_t^{(\mathrm{ME})})
\in\mathbb R^{4\times35}.
$

\paragraph{Feature construction, training and evaluation.}
For AMI, we use a practical adaptation of \textsc{RoleTrace}. We construct the text component $u_t^{(g)}$ from the manual transcript. Participant text within each window is first embedded into 384 dimensions and then projected to 32 dimensions by PCA. The PCA projection is fitted separately within each fold using only the fit partition and is shared across the four participant groups. The three activity features in Eq.~\ref{eq:ami_group_observation} are appended to the projected text representation, and each feature is standardized using statistics computed from the fit partition. Thus, the four groups provide a 140-dimensional observation at each time point. 
The representation model maps the grouped observation to a 32-dimensional latent state $Z_t$. The encoder uses hidden dimensions 128 and 64, followed by a 32-dimensional output. We additionally provide one observed indicator for each participant, allowing the model to distinguish training-time input masking from naturally silent participants. Temporal context is summarized by a 32-dimensional GRU over the preceding three latent states. Training combines reconstruction of the current grouped observation with prediction of future latent states and linguistic content. No next-speaker labels are used in representation training. We train with Adam for at most 60 epochs with learning rate $10^{-3}$ and batch size 256.

We define the downstream task at the same 15-second boundaries used to construct the observations. At a boundary $T$, the target is the participant associated with the first valid dialogue act whose starting time is at or after $T$. We retain a sample when both this target and a previously completed dialogue act are available. The target may therefore coincide with the previous speaker, and its starting time is not restricted to the immediately following 15-second window. This construction yields 14,028 prediction samples from the 14,513 observation windows. The four target roles, PM, ID, UI, and ME account for 4,755, 3,118, 2,947, and 3,208 samples, respectively. We use a four-class logistic regression classifier on the selected latent coordinates. Its regularization is chosen on the selection partition, and final performance is evaluated on the held-out teams.

\paragraph{Additional comparisons.}
Table~\ref{tab:ami-fixed-input} collects the fixed-input
representation controls and other baselines from the experimental records. Their input dimensions differ from the selected latent budgets, which means they do not isolate the effect of coordinate selection at a matched budget. \textsc{PCA Observations} applies PCA directly to the original
observations and retains 32 dimensions, while \textsc{Full Observations} uses the complete 140-dimensional input without latent compression. We also include \textsc{SVM} and \textsc{XGBoost}~\citep{10.1145/2939672.2939785} as conventional supervised predictors, together with \textsc{TS2Vec}~\citep{yue2021ts}, which learns a general-purpose temporal representation through contrastive learning. These methods use different fixed input dimensions and therefore serve as supplementary references to the matched-budget comparisons in Table~\ref{tab:ami-c-comparison}. Over five seeds, the full 32-dimensional \textbf{Role $\mathbb C$} representation achieves $\bm{51.500\pm0.135}\%$ accuracy, exceeding the strongest of these references.

\begin{table}[t]
    \centering
    \caption{
        Fixed-input reference results on AMI.
        Dimensions refer to the inputs supplied to the respective predictors. These rows are separate from the matched latent-budget comparisons.
    }
    \label{tab:ami-fixed-input}

    \begingroup
    \small
    \setlength{\tabcolsep}{6pt}
    \renewcommand{\arraystretch}{1.08}

    \begin{tabular*}{\linewidth}{
        @{\extracolsep{\fill}}lcc@{}
    }
        \toprule
        Method / Representation
        & Input dimension
        & Accuracy (\%) \\
        \midrule


        PCA observations
        & $32$
        & $50.969\pm0.000$ \\

        Full observations
        & $140$
        & $50.657\pm0.166$ \\



        SVM
        & $188$
        & $51.240\pm0.143$ \\

        XGBoost
        & $188$
        & $50.783\pm0.117$ \\

        TS2Vec
        & $32$
        & $50.623\pm0.096$ \\

        Role $\mathbb C$
        & $32$
        & \bm{$51.500\pm0.135$} \\

        \bottomrule
    \end{tabular*}

    \endgroup
\end{table}

\paragraph{AMI related work.} It is noteworthy that AMI has supported a broad range of multi-party interaction tasks. Earlier work used linguistic and temporal context features for decision detection and social role recognition~\citep{frampton-etal-2009-real, 6289057, Sapru_ICASSP_2014}. Subsequent methods for meeting summarization increasingly incorporated richer interaction structure, including multimodal hierarchical attention~\citep{li-etal-2019-keep}, hierarchical speaker-aware representations~\citep{zhu-etal-2020-hierarchical} and discourse graphs~\citep{ijcai2021p524}. More recent studies further emphasize selective and structured use of meeting information, including extraction of essential content~\citep{tan-etal-2023-reconstruct}, content selection~\citep{deng2023aspect} and the modeling of speaker dynamics~\citep{gruttadauria2024online,hilgert-niehues-2025-next,li2026rethinking}. However, the basic structural units remain speakers, utterances, discourse relations, or content spans. Our formulation targets a different structural object. We treat participant roles as observation groups and identify the group-specific expression and temporal propagation roles of the latent information factors underlying the interaction.

\end{document}